%% file: ms.tex
\documentclass{article}
\pdfoutput=1

\usepackage[final]{corl_2019}

\usepackage{amsmath}
\input{math_commands.tex}

\usepackage{hyperref}
\hypersetup{
    colorlinks=true,
    linkcolor=blue,
    filecolor=magenta,      
    urlcolor=cyan,
    citecolor=blue
}

\usepackage{dsfont}
\usepackage{graphicx}
\usepackage{subcaption}
\usepackage{multirow}
\usepackage{chngpage}
\usepackage{listings}
\usepackage[frozencache,cachedir=.]{minted}

\usepackage{mdframed}
\newmdtheoremenv{hyp}{Hypothesis}

\newcommand{\tpm}[2]{$#1 \pm #2$}
\newcommand{\myKL}[2]{\textnormal{KL}\left(#1||#2\right)}

\newcommand{\rectbra}[1]{\left[#1\right]}
\newcommand{\parbra}[1]{\left(#1\right)}
\newcommand{\expmarg}{\rho^{\textnormal{exp}}(s,a)}
\newcommand{\polmarg}{\rho^\pi(s,a)}
\newcommand{\expmargt}{\rho^{\textnormal{exp}}(s_t,a_t)}
\newcommand{\polmargt}{\rho^\pi(s_t,a_t)}

\newcommand{\LOG}{\textnormal{log }}
\newcommand{\Hcausal}{\mathcal{H}^{\textnormal{causal}}}

\newcommand{\targetstatemarginal}{\rho^{\textnormal{target}}(s)}
\newcommand{\policystatemarginal}{\rho^{\pi}(s)}

\title{
A Divergence Minimization Perspective \\ on Imitation Learning Methods
}

\author{
  Seyed Kamyar Seyed Ghasemipour\\
  University of Toronto, Vector Institute \\
  \texttt{kamyar@cs.toronto.edu} \\
  \And
  Richard Zemel \\
 University of Toronto, Vector Institute\\
\texttt{zemel@cs.toronto.edu} \\
  \AND
  Shixiang Gu \\
  Google Brain\\
\texttt{shanegu@google.com} \\
}

\begin{document}
\maketitle

%===============================================================================

\begin{abstract}
%{\color{magenta} needs work}
In many settings, it is desirable to learn decision-making and control policies through learning or bootstrapping from expert demonstrations. The most common approaches under this Imitation Learning (IL) framework are Behavioural Cloning (BC), and Inverse Reinforcement Learning (IRL). Recent methods for IRL have demonstrated the capacity to learn effective policies with access to a very limited set of demonstrations, a scenario in which BC methods often fail. Unfortunately, due to multiple factors of variation, directly comparing these methods does not provide adequate intuition for understanding this difference in performance. In this work, we present a unified probabilistic perspective on IL algorithms based on divergence minimization. We present $f$-MAX, an $f$-divergence generalization of AIRL \citep{DBLP:journals/corr/abs-1710-11248}, a state-of-the-art IRL method. $f$-MAX enables us to relate prior IRL methods such as GAIL~\citep{DBLP:journals/corr/HoE16} and AIRL~\citep{DBLP:journals/corr/abs-1710-11248}, and understand their algorithmic properties. Through the lens of divergence minimization we tease apart the differences between BC and successful IRL approaches, and empirically evaluate these nuances on simulated high-dimensional continuous control domains. Our findings conclusively identify that IRL's state-marginal matching objective contributes most to its superior performance. Lastly, we apply our new understanding of IL method to the problem of state-marginal matching, where we demonstrate that in simulated arm pushing environments we can teach agents a diverse range of behaviours using simply hand-specified state distributions and no reward functions or expert demonstrations. For datasets and reproducing results please refer to \url{https://github.com/KamyarGh/rl_swiss/blob/master/reproducing/fmax_paper.md}.

\end{abstract}


%===============================================================================

\input{intro}
\input{related_work}
\input{background}
\input{fmax}
\input{relations}
\input{fairl}
\input{reward_intuition}

\input{irl_vs_bc_exps}

\input{state_marginal_matching}
\input{conclusion}
\clearpage

% The acknowledgments are automatically included only in the final version of the paper.
% \acknowledgments{}
{
\footnotesize
\bibliography{bibrefs}  % .bib
}
\clearpage
\appendix
\input{appendix}

\end{document}

%% file: math_commands.tex
\usepackage{amsmath,amsfonts,bm}

\def\eqref#1{equation~\ref{#1}}
\def\1{\bm{1}}

\DeclareMathAlphabet{\mathsfit}{\encodingdefault}{\sfdefault}{m}{sl}
\SetMathAlphabet{\mathsfit}{bold}{\encodingdefault}{\sfdefault}{bx}{n}

\DeclareMathOperator*{\argmax}{arg\,max}
\DeclareMathOperator*{\argmin}{arg\,min}

%% file: intro.tex
\section{Introduction}
%SG0706: the intro needs some rewriting. Currently OUR contribution and problem statement is way too short (all in last paragraph). You can cut the whole sections or quite a few sentences from: (1) why lfd is important (don't need a paragraph), (2) why max ent irl is introduced (it's not directly relevant for what we are proposing). You at most should jump into YOUR problem within two paragraphs of intro. 
%SG0706: add enough citations. Intro is where you summarize you know enough about the field and then proposing novel problem and/or solutions. Adding citations around help make that point. 

    Modern advances in reinforcement learning (RL) aim to alleviate the need for hand-engineered decision-making and control algorithms by designing general purpose methods that learn to optimize provided reward functions. In many cases however, it is either too challenging to optimize a given reward (e.g. due to sparsity of signal), or it is simply impossible to design a reward function that captures the intricate details of desired outcomes~\citep{atkeson1997robot,schaal1997learning,argall2009survey,pastor2009learning,zhu2018reinforcement,rajeswaran2017learning}. One approach to overcoming such hurdles is Imitation Learning (IL) or Learning from Demonstrations (LfD)~\citep{atkeson1997robot,schaal1997learning,abbeel2004apprenticeship}
    where algorithms are provided with expert demonstrations of how to accomplish desired tasks.
    
    The most common approaches in IL framework are Behavioural Cloning (BC) and Inverse Reinforcement Learning (IRL) \citep{russell1998learning, ng2000algorithms}. 
    %%(done) SG.03.19: reference DAgger somewhere in this paragraph? cite Pieter Abbeel's paper and GAIL.
    In standard BC, learning from demonstrations is treated as a supervised learning problem and policies are trained to regress expert actions from a dataset of expert demonstrations.
    % Other forms of Behaviour Cloning, such as DAgger \citep{ross2011reduction}, consider how to make use of an expert in a more optimal fashion.
    On the other hand, in IRL the aim is to infer the reward function of the expert, and subsequently train a policy to optimize this reward. The motivation for IRL stems from the intuition that the reward function is the most concise and portable representation of a task \citep{ng2000algorithms, abbeel2004apprenticeship}.

    Recent ``adversarial" IRL methods \citep{DBLP:journals/corr/HoE16, finn2016connection, DBLP:journals/corr/abs-1710-11248} have shown tremendous success in benchmarks for continuous control \citep{brockman2016openai}; these methods outperform Behaviour Cloning by a wide margin, particularly in the low data regime where a very limited number of expert trajectories are available.
    However, it is not immediately clear why adversarial IRL methods outperform BC, since at optimality both methods exactly recover the expert policy. This question motivates the work presented here.
    %%(done) SG.03.19: include reference to f-divergence and discuss mode-seeking, mode-covering behaviors.
    
    The contribution of this work are as follows. Drawing upon the literature on $f$-divergences \citep{lin1991divergence, nowozin2016f}, we begin by presenting $f$-MAX, an algorithm for Max-Ent IRL. 
    We demonstrate how $f$-MAX generalizes AIRL \citep{DBLP:journals/corr/abs-1710-11248}, and provides new intuition for what similar algorithms accomplish.
    % RZ: I'd leave this next sentence out
    %Specifically we demonstrate that the objective in
    %%(done) SG.03.19: use "AIRL~\citep{...}" instead of just Fu.
    %AIRL is equivalent to minimizing the reverse KL divergence between the joint state-action %marginal distribution of the policy and that of the expert.
    % Additionally, we demonstrate that $f$-MAX, and by inheritance AIRL, is a subset of the cost-regularized Max-Ent IRL framework laid out by \cite{DBLP:journals/corr/HoE16}.
    %RZ: This next sentence should go at the end - as in the paper, the hypotheses and
    %evaluations after the analysis
    From our findings we generate hypotheses for why Max-Ent IRL methods outperform BC, and empirically evaluate them in continuous control benchmarks.
    Through this process we gain a unified understanding of Imitation Learning methods from a divergence minimization perspective.
    To demonstrate the versatility of this perspective, we present a new approach to the recently proposed problem of state-marginal matching \citep{lee2019efficient} and show that we can train policies for a diverse range of behaviours using no reward functions or expert demonstrations.
    
    % To tease apart the differences between standard BC and AIRL, we also address the degeneracy of $f$-MAX in a special case, and provide a one-line modification of AIRL, named FAIRL, which minimizes the \emph{forward} KL divergence between the joint state-action marginal of the expert and the policy. We discuss how our findings may relate to common observations \citep{bishop2006pattern} regarding the mode-covering/mode-seeking behaviour of the different KL divergence directions.

    % To our surprise, 
    % %%(done) SG.03.19: learns policy as well as AIRL~\cite{}. 
    % while this method learns a policy that performs comparably to AIRL\citep{DBLP:journals/corr/abs-1710-11248}, 
    % %%(done) SG.03.19: "it can do so.."
    % it does so with \emph{significantly} less interactions with the environment and less number of training epochs.
    

%% file: related_work.tex
\section{Related Work}
    The connections between RL and divergence minimization have long been studied in the rich prior literature of control as probabilistic inference~\cite{todorov2007linearly,toussaint2009robot,peters2010relative,kappen2012optimal,ziebart2008maximum,ziebart2010modeling,haarnoja2018soft}. Specifically, they have shown that optimal control under entropy regularization can be viewed as approximate inference on a graphical model, or equivalently minimizing reverse KL divergence between reward-weighted trajectory and policy trajectory distributions~\citep{kappen2012optimal,levine2018reinforcement}. 
    Building on such intuitions, a number of work extended RL algorithms based on picking another divergence metric, such as forward KL~\citep{peters2007reinforcement,norouzi2016reward}, and demonstrated substantially improved empirical performances in certain situations.
    %  or Stein discrepancy~\citep{liu2017stein}
    %While the Maximum-Entropy (Max-Ent) RL framework can be seen a reverse KL minimization, forward KL variants have also been shown to perform well in certain cases~\citep{peters2007reinforcement,norouzi2016reward}. 
    Our work draws significant inspirations from these prior works in RL and aims to provide a probabilistic perspective in Imitation Learning (IL). 
    
    In the field of robotics, imitation learning (IL), or bootstrapping from IL, has often been the method of choice over RL due to difficulty in exploration and scarcity of data~\citep{atkeson1997robot,schaal1997learning,argall2009survey,pastor2009learning}. While Behavioural Cloning (BC) is the most widely used IL algorithm due to the simplicity of its objective, it suffers from the problem of covariate shift between train and test time.
    %, a common problem in supervised sequence prediction tasks~\citep{bengio2015scheduled}. 
    Methods such as DAgger~\citep{ross2011reduction} and Dart~\citep{laskey2017iterative} aim to relieve this mismatch, yet assume interactive access to expert policies.
    %Another important approach to Imitation Learning is 
    
    Inverse Reinforcement Learning (IRL) algorithms have shown promising results in challenging continuous control problems~\citep{coates2008learning,abbeel2010autonomous, DBLP:journals/corr/HoE16,DBLP:journals/corr/abs-1710-11248, zhu2018reinforcement}, outperforming BC. Similarly to RL, the connections between IRL and divergence minimization have long been alluded. Early works in IRL operated by matching feature expectations or moments~\citep{abbeel2004apprenticeship} between policies and experts, a popular approach in distribution matching~\citep{dziugaite2015training,li2015generative}. Furthermore, Maximum Entropy (Max-Ent) IRL~\citep{ziebart2008maximum,ziebart2010modeling} --- an IRL framework that addresses degeneracies of the original IRL formulation~\citep{russell1998learning, ng2000algorithms} --- is explicitly formulated as an energy-based modeling problem. Recent scalable approaches to Max-Ent IRL~\citep{DBLP:journals/corr/HoE16,DBLP:journals/corr/abs-1710-11248,finn2016connection}, motivated by adversarial approaches to generative modeling~\cite{goodfellow2014generative}, demonstrate additional connections to distribution matching. Our work generalizes the objective proposed in~\citep{DBLP:journals/corr/abs-1710-11248,finn2016connection} based on recent insights from generative modeling~\citep{nowozin2016f}, and further provides a unified perspective for viewing common IL algorithms. Concurrent to our work,~\citet{ke2019imitation} also present a unifying probabilistic perspective on IL; however, their empirical experiments solely focus on grid world domains, while our work provides comparative results on high-dimensional continuous control environments and also evaluates the effectiveness of IL algorithms for state marginal matching~\citep{lee2019efficient}.  

%% file: background.tex
\section{Background}
    Consider a Markov Decision Process (MDP) represented as a tuple $(\mathcal{S}, \mathcal{A}, \mathcal{P}, r, \rho_0, \gamma)$ with state-space $\mathcal{S}$, action-space $\mathcal{A}$, dynamics $\mathcal{P}:\mathcal{S}\times\mathcal{A}\times\mathcal{S}\rightarrow [0,1]$, reward function $r(s,a)$, initial state distribution $\rho_0$, and discount factor $\gamma \in (0,1)$. Throughout this work we will denote the marginal state-action and marginal state distribution of a policy by $\polmarg$ and $\rho^{\pi}(s)$ respectively.\footnote{
        Intuitively, the marginal distributions of a policy are obtained by generating infinitely many trajectories (i.e. finite or infinite horizon episodes) with a the given policy and computing the frequency of $(s,a)$ or $(s)$.
    } 
    
    % \subsection{Adversarial Methods for IRL}
    \paragraph{Adversarial Methods for IRL}
    Instead of recovering the reward function and policy, recent successful methods in Maximum-Entropy IRL (Max-Ent IRL) aim to directly recover the policy resulting from the full process.
    
    % Since such methods only recover the policy, it would be more accurate to refer to them as Imitation Learning algorithms. However, to avoid confusion with Behaviour Cloning methods, in this work we will refer to them as \emph{direct} methods for Max-Ent IRL.
        
    \paragraph{GAIL: Generative Adversarial Imitation Learning}
        \label{sec:gail}
        Before describing the work of \cite{DBLP:journals/corr/HoE16}, we establish the definition of causal entropy $\Hcausal(\pi) := \mathds{E}_{\rho^\pi(s)} \rectbra{\LOG \pi(a|s)}$ \citep{ziebart2010modeling, bloem2014infinite}. Intuitively, causal entropy can be thought of as the ``amount of options" the policy has in each state, in expectation.
    
        Let $\mathcal{C}$ denote a class of cost functions (negative reward functions). Furthermore, let $\expmarg, \polmarg$ denote the state-action marginal distributions of the expert and student policy respectively. \citet{DBLP:journals/corr/HoE16} begin with a regularized Max-Ent IRL objective,
            \begin{align}
                \textnormal{IRL}_{\psi}(\pi^{\textnormal{exp}}) := &\argmax_{c \in \mathcal{C}} - \psi(c) + \parbra{\min_\pi -\Hcausal(\pi) + \mathds{E}_{\polmarg} \rectbra{c(s,a)}}
                - \mathds{E}_{\expmarg} \rectbra{c(s,a)}
                \label{eq:reg_irl}
            \end{align}
        where $\psi: \mathcal{C} \rightarrow \mathds{R}$ is a convex regularization function on the space of cost functions, and $\textnormal{IRL}_\psi(\pi^\textnormal{exp})$ returns the optimal cost function given the expert and choice of regularization.
        % Also, while not immediately clear, note that $\min_\pi -\Hcausal(\pi) + \mathds{E}_\pi \rectbra{c(s,a)}$ is the Max-Ent RL objective given cost function $c(s,a)$.
        Let $\textnormal{RL}(c) := \argmin_\pi -\Hcausal(\pi) + \mathds{E}_\pi \rectbra{c(s,a)}$, be a function that returns the optimal Max-Ent policy given cost $c(s,a)$. \citet{DBLP:journals/corr/HoE16} show that
            \begin{align}
                \textnormal{RL} \circ \textnormal{IRL}_\psi (\pi^{\textnormal{exp}}) = \argmin_\pi -\Hcausal(\pi) + \psi^*\parbra{\polmarg - \expmarg}
                \label{eq:gail_general_objective}
            \end{align}
        where $\psi^*$ denotes the convex conjugate of $\psi$. This tells us that if we were to find the cost function $c(s,a)$ using the regularized Max-Ent IRL objective \ref{eq:reg_irl}, and subsequently find the optimal Max-Ent policy for this cost, we would arrive at the same policy had we directly optimized objective \ref{eq:gail_general_objective}.% by searching for the policy.
        
        % {\color{magenta} should probably also say something about them showing Max-Ent IRL is equivalent to distribution matching. But also, this can be shown much more trivially than how they showed it.}
        
        Directly optimizing Equation \ref{eq:gail_general_objective} is challenging for many choices of $\psi$.
        % Interestingly however, \cite{DBLP:journals/corr/HoE16} show that for any symmetric $f$-divergences \citep{lin1991divergence}, there exists a choice of $\psi$ such that equation \ref{eq:gail_general_objective} is equivalent to $\textnormal{RL} \circ \textnormal{IRL}_\psi (\pi^{\textnormal{exp}}) = \argmin_\pi \Hcausal(\pi) + D_f\parbra{\polmarg || \expmarg}$. In such settings, due to a close connection between binary classifiers and symmetric $f$-divergences \citep{nguyen2009surrogate}, efficient algorithms can be formed.
        A special case
        % for Jensen-Shannon divergence
        leads to the successful method dubbed Generative Adversarial Imitation Learning (GAIL). As before, let $\expmarg, \polmarg$ denote the state-action marginal distributions of the expert and student policy respectively. Let $D(s,a): \mathcal{S}\times\mathcal{A} \rightarrow [0,1]$ be a binary classifier - often referred to as the discriminator - for identifying positive samples (sampled from $\expmarg$) from negative samples (sampled from $\polmarg$). Using RL, the student policy is trained to maximize $\mathds{E}_{\tau\sim\pi}\rectbra{\sum_t \LOG D(s_t,a_t)} - \lambda \Hcausal(\pi)$, where $\lambda$ is a hyperparameter. The training procedure alternates between optimizing the discriminator and updating the policy;
        %. As noted, it is shown that this training 
        this procedure minimizes the Jensen-Shannon divergence between $\expmarg$ and $\polmarg$ \citep{DBLP:journals/corr/HoE16}.
        
        % Further connections between this work and ours are discussed in section \ref{sec:more_gail_connections}.
        
        %%(done) SG.03.19: can we discuss difference between imitation learning (doesn't need to recover reward) and inverse RL somewhere? 
    
    \paragraph{AIRL: Adversarial Inverse Reinforcement Learning \label{sec:airl}}
        Subsequent to the advent of GAIL \citep{DBLP:journals/corr/HoE16}, \citet{finn2016connection} present a theoretical discussion relating Generative Adversarial Networks (GANs) \citep{goodfellow2014generative}, IRL, and energy-based models. They demonstrate how an adversarial training approach could recover the Max-Ent reward function and simultaneously train the Max-Ent policy corresponding to that reward.
        Building on this discussion, \citet{DBLP:journals/corr/abs-1710-11248} present a practical implementation of this method, named Adversarial IRL (AIRL). The main difference between AIRL \citep{DBLP:journals/corr/abs-1710-11248} and GAIL \citep{DBLP:journals/corr/HoE16} arises from the objective used to train the policy: in AIRL, the policy optimizes $\mathds{E}_{\tau\sim\pi}\rectbra{\sum_t \LOG D(s_t,a_t) - \LOG (1-D(s_t,a_t))}$. \citet{DBLP:journals/corr/abs-1710-11248} also present additional contributions regarding the recovery of the expert reward function which is not directly relevant to this work.
        
    \paragraph{Performance With Respect to BC}
        Methods such as GAIL and AIRL have demonstrated significant performance gains compared to Behavioural Cloning. In particular, in standard Mujoco benchmarks \citep{todorov2012mujoco, brockman2016openai}, adversarial methods for Max-Ent IRL achieve strong performance using a \emph{very limited} amount of expert demonstrations, an important failure scenario for standard BC.

    % % ----------------------------------------------------------
    % \subsection{Standard But Important Observation}
    %     \label{sec:observation}

    %     %%SG.03.19: i'm not sure how this is new. isn't this standard observation in abbeel's paper or GAIL? the notation here is very informal. I suggest removing this section. 
    %     \paragraph{Statement:} If in Max-Ent IRL all we care about is to recover the optimal policy, this problem is equivalent to matching $\expmarg$ and $\polmarg$\footnote{This statement is motivate by intuitions from \citep{DBLP:journals/corr/HoE16}}.
        
    %     \paragraph{Proof:}
    %     The reverse direction of the proof is trivial due to the equivalence between policies and state-action marginals. To see the forward direction, we recall that in the Max-Ent \emph{Reinforcement Learning} setup, for any reward function there exists a unique optimal policy. Note that trajectories from the optimal policy follow the distribution: $\tau \sim \frac{1}{Z}\textnormal{exp}(R(\tau))$. Hence, if two policies $\pi_1$ and $\pi_2$ are optimal, they will induce the same distribution over trajectories, and therefore induce the same state-action marginal distributions as well. This then implies that the two policies are identical since, $\pi_1(a|s) = \frac{\rho^{\pi_1}(s,a)}{\int_\mathcal{A} \rho^{\pi_1}(s,a)} = \frac{\rho^{\pi_2}(s,a)}{\int_\mathcal{A} \rho^{\pi_2}(s,a)} = \pi_2(a|s)$.
    
    % ----------------------------------------------------------
    % \subsection{$f$-GAN}
    \paragraph{$f$-GAN}
        \label{sec:f_div}
        %%(done) SG.03.19: this section comes out very randomly. add f-divergence ref in intro as suggested above + write 1-2 sentence here discussing how f-divergence relates to MaxEntIRL or GAIL. "As discussed <...>(make sure you discuss how maxent irl can be seen as minimizing KL), inverse RL can be seen as minimizing some form of divergence between \rho^exp and \rho^pi. In this section, we discuss f-divergence, which ....
        
        % \cite{DBLP:journals/corr/HoE16} demonstrate that Max-Ent IRL is the dual problem of matching $\polmarg$ to $\expmarg$; indeed as noted above, GAIL \citep{DBLP:journals/corr/HoE16} optimizes the Jensen-Shannon divergence between the two distributions. In section \ref{sec:fax} we present $f$-MAX, a method for matching $\polmarg$ to $\expmarg$ using arbitrary $f$-divergences \citep{lin1991divergence}. Hence, in this section we recall this class of statistical divergences as well as methods for using them for training generative models.
        
        Let $P, Q$ be two distributions with density functions $p, q$.
        % For any convex, lower-semicontinuous function  $f: \mathds{R}^{+} \rightarrow \mathds{R}$ a statistical divergence can be defined as: $D_f(P||Q) = \int_\chi q(x) f\left(\frac{p(x)}{q(x)}\right)$. Divergences derived in this manner are called \emph{f-divergences} and amongst many interesting divergences include the forward and reverse KL.
        % \citet{nguyen2010estimating} present a variational estimation method for $f$-divergences between arbitrary distributions P, Q. Using the notation of \citet{nowozin2016f} we can write,
            % \begin{align}
                % $D_f(P||Q) \ge \sup_{T_\omega \in \mathcal{T}}(\mathds{E}_{x \sim P}\rectbra{T_\omega(x)} - \mathds{E}_{x \sim Q}\rectbra{f^*(T_\omega(x))}) \label{eq:variational_f_div}$,
            % \end{align}
        % where $\mathcal{T}$ is an arbitrary class of functions $T_\omega: X \rightarrow \mathds{R}$, and $f^*$ is the convex conjugate of $f$. Under mild conditions equality holds between the two sides~\citep{nguyen2010estimating}.
        %%SG.03.19: can you just write "we have the equality" and no eq below?
            % \begin{align}
            %     D_f(P||Q) = \sup_{T_\omega \in \mathcal{T}}(\mathds{E}_{x \sim P}\rectbra{T_\omega(x)} - \mathds{E}_{x \sim Q}\rectbra{f^*(T_\omega(x))})
            % \end{align}
        Motivated by variational lower bounds of $f$-divergences \citep{nguyen2010estimating}, as well as Generative Adversarial Networks (GANs) \citep{goodfellow2014generative}, \citet{nowozin2016f} present an iterative optimization scheme for matching an implicit distribution\footnote{We use the term ``implicit distributions" to refer to distributions we can efficiently sample from (e.g. GAN \citep{goodfellow2016deep} generators) but do not necessarily have the densities of} Q to a fixed distribution P using any $f$-divergence. For a given $f$-divergence, the corresponding minimax optimization is,
        \begin{align}
            \min_Q \max_{T_\omega} \mathds{E}_{x \sim P}\rectbra{T_\omega(x)} - \mathds{E}_{x \sim Q}\rectbra{f^*(T_\omega(x))}
            \label{eq:fgan}
        \end{align}
        where $T_\omega: X \rightarrow \mathds{R}$, and $f^*$ is the convex conjugate of $f$. Further discussion in Appendix \ref{app:f_div}.
%        \citet{nowozin2016f} discuss practical parameterizations of $T_\omega$, but to avoid notational clutter we will use the form above.

%% file: fmax.tex
\section{$f$-MAX: $f$-Divergence Max-Ent IRL}
    \label{sec:fax}
    We begin by presenting $f$-MAX, a generalization of AIRL \citep{DBLP:journals/corr/abs-1710-11248} which provides a more intuitive interpretation of what similar algorithms accomplish.
    %%(done) SG.03.19: is this new result? if not, we can just cite. else should have proper lemma or corollary heading for section 2.3 content.
    % Given the discussion in section \ref{sec:observation},
    %%(done) SG.03.19: again, isn't imitation learning better name for such problem definition?
    % the Max-Ent IRL problem is equivalent to minimizing any statistical divergence between the expert's state-action marginal distribution, $\expmarg$, and that of the policy's, $\polmarg$. 
    %%(done) SG.03.19: remove "as foreshawdowed"
    % As noted, , we will do this by minimizing $f$-divergences between these two distributions.
    Imagine for some $f$-divergence we aim minimize $D_f\left(\expmarg||\polmarg\right)$. Using the $f$-GAN \citep{nowozin2016f} formulation this objective can be written as,
    \begin{align}
        \min_{\pi}\max_{T_\omega} \mathds{E}_{(s,a) \sim \expmarg}\rectbra{T_\omega(s,a)} - \mathds{E}_{(s,a) \sim \polmarg}\rectbra{f^*(T_\omega(s,a))} \label{eq:minimax_obj}
    \end{align}
    To optimize this objective with propose the following iterative optimization procedure,
    \begin{align}
        \label{eq:disc}
        &\max_{T_\omega} \mathds{E}_{(s,a) \sim \expmarg}\rectbra{T_\omega(s,a)} - \mathds{E}_{(s,a) \sim \polmarg}\rectbra{f^*(T_\omega(s,a))},\\
        &\quad \max_\pi \mathds{E}_{\tau \sim \pi} \rectbra{\sum_t f^*(T_\omega(s_t, a_t))} \label{eq:policy}
    \end{align}
    % Imagine, for some $f$, we aim to train a policy by optimizing the $f$-divergence $D_f\left(\expmarg||\polmarg\right)$. To do so, we propose the following iterative optimization procedure,
    % %%(done) SG.03.19: i think you can leave (7) to be in the same form as (6). and then = the form with \tau and sum_t. 
    %     \begin{align}
    %         \label{eq:disc}
    %         &\max_{T_\omega} \mathds{E}_{(s,a) \sim \expmarg}\rectbra{T_\omega(s,a)} - \mathds{E}_{(s,a) \sim \polmarg}\rectbra{f^*(T_\omega(s,a))} \\
    %         &\max_\pi \mathds{E}_{\tau \sim \pi} \rectbra{\sum_t f^*(T_\omega(s_t, a_t))} \label{eq:policy}
    %     \end{align}
    % where $f^*$ and $T_\omega$ are as defined in section \ref{sec:f_div}.
    Equation \ref{eq:disc} is the same as the inner maximization in Equation \ref{eq:minimax_obj}; this objective optimizes $T_\omega$ so that Equation \ref{eq:disc} best approximates $D_f\left(\expmarg||\polmarg\right)$. On the other hand, using the identities in appendix \ref{ap:deriv} we have that up to a multiplicative constant,
    $
        \mathds{E}_{\tau \sim \pi} \rectbra{\sum_t f^*(T_\omega^\pi(s_t, a_t))} \propto
        % \int_{S,A} \polmarg \rectbra{f^*(T_\omega^\pi(s, a))}\\
        \mathds{E}_{(s,a) \sim \polmarg}\rectbra{f^*(T_\omega^\pi(s,a))}
    $.
    This implies that the policy objective (Equation \ref{eq:policy}) is equivalent to minimizing Equation \ref{eq:disc} with respect to $\pi$.
    With an identical proof as in \citet[Proposition~2]{goodfellow2014generative}, if in each iteration the optimal $T_\omega$ is found, the described optimization procedure converges to the global optimum where the policy's state-action distribution matches that of the expert.% This is equivalent to iteratively computing $D_f\left(\expmarg||\polmarg\right)$ and optimizing the policy to minimize it.

    \subsection{Corollary: A Simple Derivation and Intuition for AIRL}
    % \paragraph{Corollary: A Simple Derivation and Intuition for AIRL}
        \label{sec:corollary}
        Choosing $f(u) := -\LOG u$ leads to $D_f(\expmarg || \polmarg) = \myKL{\polmarg}{\expmarg}$. This divergence is commonly referred to as the ``reverse" KL divergence. In this setting we have, $f^*(t) = -1 - \LOG(-t)$, and $T_\omega^\pi(s,a) = -\frac{\polmarg}{\expmarg}$ \citep{nowozin2016f}. As we demonstrate in Appendix \ref{app:airl_is_rev_kl}, in this setting $f$-MAX is equivalent to AIRL \citep{DBLP:journals/corr/abs-1710-11248}, meaning that AIRL is solving the Max-Ent IRL problem by minimizing the reverse KL divergence.
        
    \subsection{Relation to Cost-Regularized Max-Ent IRL}
    % \paragraph{Relation to Cost-Regularized Max-Ent IRL}
        \label{sec:more_gail_connections}
        As discussed above, \citet{DBLP:journals/corr/HoE16} present a class of methods for Max-Ent IRL that directly retrieve the expert policy without explicitly finding the reward function of the expert (sec. \ref{sec:gail}). Additionally, they present practical approaches for minimizing any \emph{symmetric}\footnote{We call an $f$-divergence symmetric if for any P,Q we have $D_f(P||Q) = D_f(Q||P)$} $f$-divergence between $\expmarg$ and $\polmarg$. Choosing the symmetric $f$-divergence to be the Jensen-Shannon divergence leads to the successful special case, GAIL (sec \ref{sec:gail}).
        
        %Surprisingly, 
        We can show that $f$-MAX is a subset of the cost-regularized Max-Ent IRL framework of \citet{DBLP:journals/corr/HoE16}.
        % Recall the following equations from this framework,
        %     \begin{align}
        %                 &\textnormal{IRL}_{\psi}(\pi^{\textnormal{exp}}) := \argmax_{c \in \mathcal{C}} - \psi(c)
        %                 + \parbra{\min_\pi -\Hcausal(\pi) + \mathds{E}_{\polmarg} \rectbra{c(s,a)}}
        %                 - \mathds{E}_{\expmarg} \rectbra{c(s,a)}\\
        %                 &\textnormal{RL} \circ \textnormal{IRL}_\psi (\pi^{\textnormal{exp}}) = \argmin_\pi -\Hcausal(\pi) + \psi^*\parbra{\polmarg - \expmarg}
        %         \label{eq:recall_reg_airl}
        %     \end{align}
        % where $\psi(c): \mathcal{C} \rightarrow \mathds{R}$ was a closed, proper, and convex regularization function on the space of cost function, and $\psi^*$ its convex conjugate.
        For a given $f$-divergence, choosing $\psi(c) := \mathds{E}_{\expmarg}\rectbra{f^*(c(s,a)) - c(s,a)}$ we obtain\footnote{Full derivations can be found in Appendix \ref{ap:gail_relation}},
        \begin{align}
            &\psi^*_f(\polmarg - \expmarg) = D_f \parbra{\polmarg || \expmarg}\\
            &\textnormal{RL} \circ \textnormal{IRL}_\psi (\pi^{\textnormal{exp}}) = \argmin_\pi -\Hcausal(\pi) + D_f \parbra{\polmarg || \expmarg}
        \end{align}
        Typically, the causal entropy term is considered a policy regularizer, and is weighted by $0 \le \lambda \le 1$. Therefore, modulo the term $\Hcausal(\pi)$, our derivations show that $f$-MAX, and by inheritance AIRL \citep{DBLP:journals/corr/abs-1710-11248}, all fall under the cost-regularized Max-Ent IRL framework of \citet{DBLP:journals/corr/HoE16}!
        % Furthermore, this demonstrates that the reward function of the expert can be recovered as $-T^*_\omega(s,a)$.
        

%% file: relations.tex
\section{Understanding the Relation Among Imitation Learning Methods}
    
    {\renewcommand{\arraystretch}{1.5}
        \begin{table*}[t]
            \begin{adjustwidth}{-1in}{-1in}
            \centering
            \begin{tabular}{|c|l|}
                \hline
                \textbf{Method} & \multicolumn{1}{c|}{\textbf{Optimized Objective (Minimization)}}\\ \hline
                Standard Behavioural Cloning & $\mathds{E}_{\rho^{\textnormal{exp}}(s)}\rectbra{\myKL{\pi^{\textnormal{exp}}(a|s)}{\pi(a|s)}} = -\mathds{E}_{\expmarg} \rectbra{\LOG \pi(a|s)}+C$\\
                DAgger \citep{ross2011reduction} & $\mathds{E}_{\rho^{\textnormal{agg}_{1:n}}(s)}\rectbra{\myKL{\pi^{\textnormal{exp}}(a|s)}{\pi(a|s)}}$ at iteration $n+1$\\
                AIRL \citep{DBLP:journals/corr/abs-1710-11248} & $\textnormal{KL}(\polmarg||\expmarg) = -\mathds{E}_{\polmarg} \rectbra{\LOG \expmarg} - \mathcal{H}(\polmarg)$\\
                GAIL \citep{DBLP:journals/corr/HoE16} & $D_{\textnormal{JS}}(\polmarg || \expmarg) - \lambda \mathcal{H}^{\textnormal{causal}}(\pi)$ \\
                FAIRL (this work, section \ref{sec:fairl}) & $\textnormal{KL}(\expmarg||\polmarg) = -\mathds{E}_{\expmarg} \rectbra{\LOG \polmarg} - \mathcal{H}(\expmarg)$\\
                symmetric $f$-div~\citep{DBLP:journals/corr/HoE16} & $D_{f\textnormal{-symm}}(\polmarg || \expmarg) - \lambda \mathcal{H}^{\textnormal{causal}}(\pi)$ \\
                $f$-MAX (this work, section \ref{sec:fax}) & $D_{f}(\polmarg || \expmarg)$ \\
                \hline
            \end{tabular}
            \end{adjustwidth}
            \caption{
            {
                \small
                The objective function for various imitation learning algorithms, written in a common form as the minimization of statistical divergences. $\mathcal{H}(\cdot)$ denotes entropy, $\mathcal{H}^{\textnormal{causal}}(\pi)$ denotes the causal entropy of the policy \citep{ziebart2010modeling, DBLP:journals/corr/HoE16}, and $\lambda$ is a hyperparameter. JS denotes the Jensen-Shannon divergence and $D_f$ indicates any $f$-divergence. For DAgger, we are showing the objective for the simplest form of the algorithm, where $\pi^{(i)}$ is the policy obtained at iteration $i$, $\pi^{(1)}$ is the expert, and $\rho^{\textnormal{agg}_{1:n}}(s) = \frac{1}{n} \sum_{i=1}^{n} \rho^{\pi^{(i)}}(s)$.}
                % \vspace{-0.5cm}
            }
            \label{tab:comparisons}
        \end{table*}
    }

    %%(done) SG.03.19: this is far too late in appearing. in the background section for maxent irl, you should add bc discussion and write both of them in the form of divergence minimization. plus. you should add a table where it summarizes maxent irl, airl, bc, yours, forward kl case (fairl). i think that table was one of the best parts of contribution.  
    Given results derived in the prior section we can now begin to populate Table \ref{tab:comparisons}, writing various IL algorithms in a common form, as the minimization of some statistical divergence between $\expmarg$ and $\polmarg$. In BC we minimize $\mathds{E}_{\rho^{\textnormal{exp}}(s)}\rectbra{\myKL{\pi^{\textnormal{exp}}(a|s)}{\pi(a|s)}}$\footnote{Since $\mathds{E}_{\rho^{\textnormal{exp}}(s)}\rectbra{\myKL{\rho^{\textnormal{exp}}(a|s)}{\rho^{\pi}(a|s)}} = -\mathds{E}_{\expmarg}\rectbra{\LOG \rho^\pi(a|s)} - \mathcal{H}^{\textnormal{exp}}(s,a)$ and $\mathcal{H}^{\textnormal{exp}}(s,a)$ is constant w.r.t. the policy ($\mathcal{H}^{\textnormal{exp}}(s,a)$ is the entropy of $\expmarg$)}. On the other hand, the corollary in section \ref{sec:corollary} demonstrates that AIRL \citep{DBLP:journals/corr/abs-1710-11248} minimizes $\myKL{\polmarg}{\expmarg}$, while GAIL \citep{DBLP:journals/corr/HoE16} optimizes $D_{JS}(\expmarg||\polmarg) - \lambda \mathcal{H}^{\textnormal{causal}}(\pi)$. Hence, there are two ways in which adversarial IRL methods differ from Behavioural Cloning. First, in standard BC the policy is optimized to match the conditional distribution $\pi^{\textnormal{exp}}(a|s)$, whereas in the other two the policy is explicitly encouraged to match the marginal state distributions as well. Second, in BC we make use of the forward KL divergence, whereas AIRL and GAIL use divergences that exhibit more mode-seeking behaviour. These observations allow us to generate the following two hypotheses about why IRL methods outperform BC, particularly in the low-data regime,
    
    \begin{hyp}
        \label{hyp:1}
        In common MDPs of interest, 
        the reward function depends more on the state than the action.
        % Hence it is plausible that matching state marginals is more useful than matching action conditional marginals.
        Hence encouraging policies to explicitly match expert state marginals is an important learning criterion.
    \end{hyp}
    \begin{hyp}
        \label{hyp:2}
        It is known that optimization using the forward KL divergence results in distributions with a mode-covering behaviour, whereas using the reverse KL results in mode-seeking behaviour \citep{bishop2006pattern}.
        In RL we care about the ``quality of trajectories",
        as measured by the likelihood under the expert distribution.
        % which in Imitation Learning is measured by the likelihood under the expert distribution.
        Therefore, being mode-seeking is more beneficial than mode-covering, particularly in the low-data regime.
        % Therefore, since in Reinforcement Learning we care about the ``quality of trajectories", being mode-seeking is more beneficial than mode-covering, particularly in the low-data regime.
    \end{hyp}
    
    In what follows, we seek to experimentally evaluate our hypotheses. To tease apart the differences between Max-Ent IRL methods and BC, we present an algorithm that optimizes the forward $\myKL{\expmarg}{\polmarg}$. We then compare its performance to Behaviour Cloning and the standard AIRL algorithm using varying amounts of expert demonstrations.
    

%% file: fairl.tex
\section{FAIRL: An Alternative Method for Forward KL}
    \label{sec:fairl}
    
    While $f$-MAX is a general algorithm, useful for most choices of $f$, it unfortunately cannot be used for the special case of forward KL, i.e. $\myKL{\expmarg}{\polmarg}$. We identify the problem in Appendix \ref{ap:forward_KL_problem} and present a separate method that optimizes this divergence.
    
    Similar to AIRL \citep{DBLP:journals/corr/abs-1710-11248}, let us have a discriminator, $D(s,a)$ whose objective is to discriminate between expert and policy state-action pairs.
    We now define the reward in Equation~\ref{eq:policy} for the policy to be,
    \begin{align}
        &h(s,a) := \LOG D(s,a) - \LOG (1 - D(s,a)), \quad r(s,a) := \textnormal{exp}(h(s,a)) \cdot (-h(s,a))%\\
       % &\textnormal{Policy Objective: }\max_\pi \mathds{E}_{\tau \sim \pi_\theta} \rectbra{\sum_t r(s_t, a_t)}
    \end{align}
    In appendix \ref{ap:FAIRL_deriv} we show that up to a multiplicative constant, $\mathds{E}_{\tau \sim \pi} \rectbra{\sum_t r(s_t, a_t)} \propto -\textnormal{KL}(\expmarg||\polmarg)$.
    This is a refreshing result since it demonstrates that we can convert the AIRL algorithm \citep{DBLP:journals/corr/abs-1710-11248} into its forward KL counterpart by simply modifying the reward function used.
    %; in AIRL (reverse KL) the reward is defined as $r(s,a) := \LOG D(s,a) - \LOG (1 - D(s,a))$, whereas for forward KL it is defined as $r(s,a) := \frac{D(s,a)}{1-D(s,a)} \cdot \LOG \frac{1 - D(s,a)}{D(s,a)}$. 
    We refer to this forward KL version of AIRL as FAIRL.

%% file: reward_intuition.tex
\section{Intuition About Different Divergence Rewards}
    \label{sec:intuition}
    \begin{figure*}[t]
        \centering
        \begin{subfigure}{0.3\textwidth}
            \centering
            \includegraphics[width=1.0\linewidth]{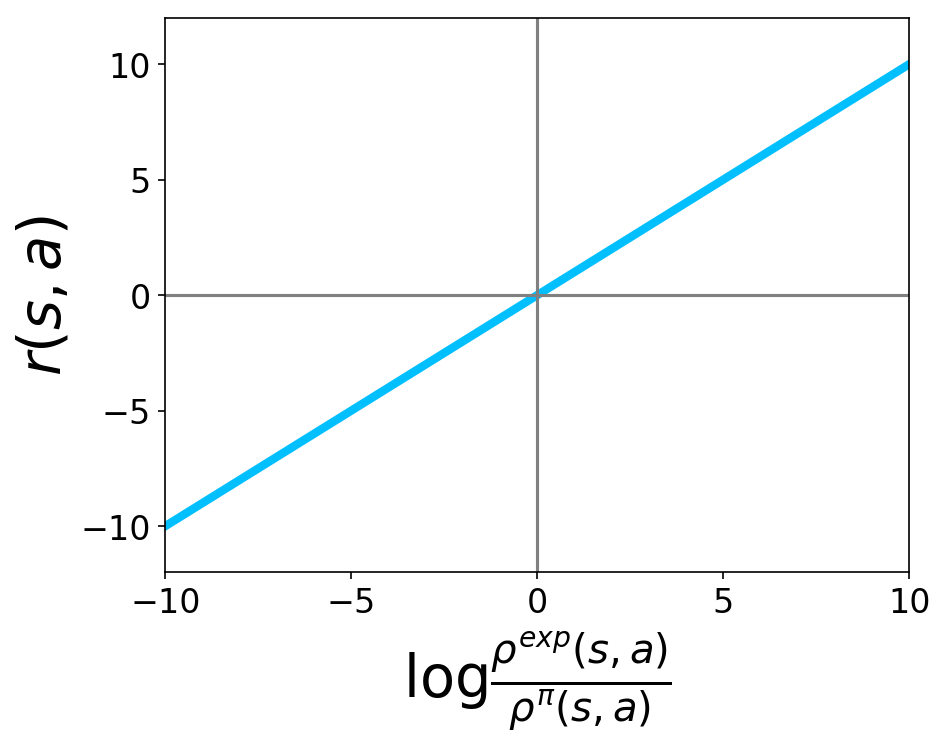}
            \caption{AIRL}
        \end{subfigure}
        \begin{subfigure}{0.3\textwidth}
            \centering
            \includegraphics[width=1.0\linewidth]{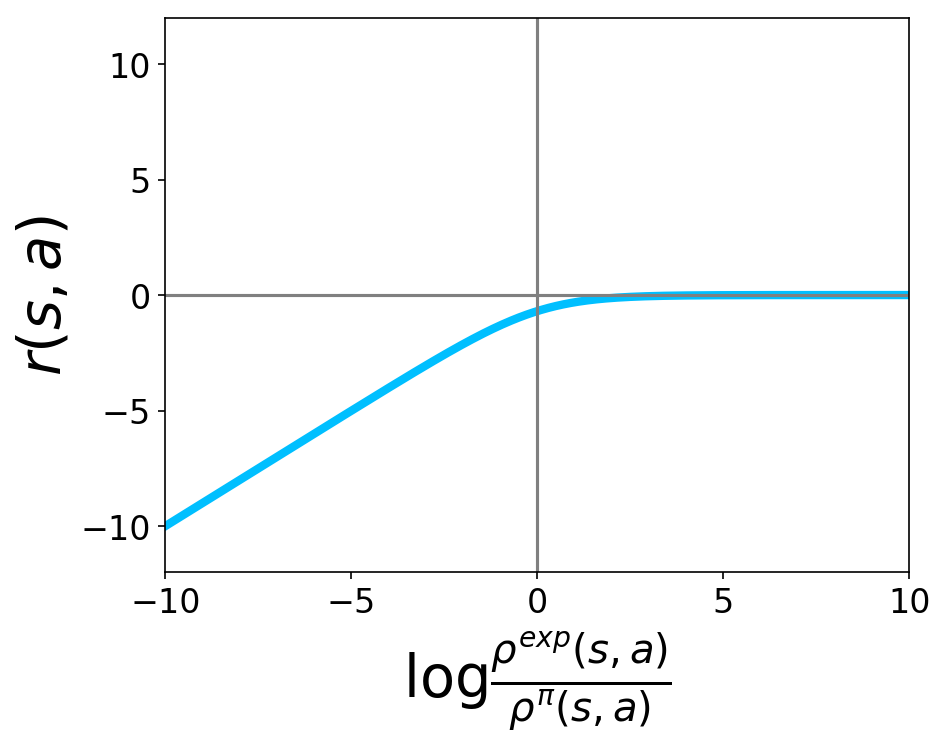}
            \caption{GAIL}
        \end{subfigure}
        \begin{subfigure}{0.3\textwidth}
            \centering
            \includegraphics[width=1.0\linewidth]{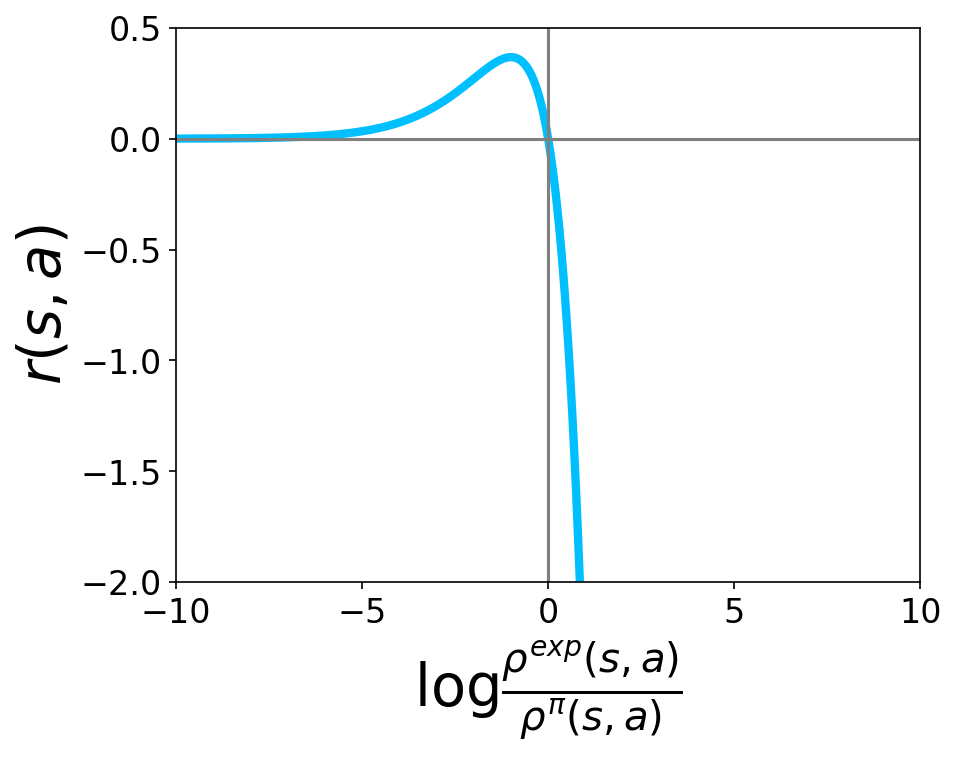}
            \caption{FAIRL}
        \end{subfigure}
        \caption{
            {
                \small
                $r(s,a)$ as the function of the logits of the optimal discriminator, $\ell^\textnormal{opt}(s,a) = \LOG \frac{\expmarg}{\polmarg}$. As a reminder, AIRL, GAIL, and FAIRL respectively correspond to the reverse KL, JS, and forward KL divergences.
            }
            % \vspace{-0.5cm}
        }
        \label{fig:kl_figs}
    \end{figure*}
    
    Let us parameterize the discriminator as $D(s,a) := \sigma(\ell(s,a))$, where $\sigma$ represents the sigmoid activation function and $\ell(s,a)$ is the logit. Rearranging equations we can write $\ell(s,a) = \LOG D(s,a) - \LOG (1 - D(s,a))$. Hence, given a policy $\pi$, for an optimal discriminator we have that the logits are equal to the log density ratio: $\ell^{\textnormal{opt}}(s,a) = \LOG \frac{\expmarg}{\polmarg}$. It is instructive to plot the reward functions of AIRL \citep{DBLP:journals/corr/abs-1710-11248}, GAIL \citep{DBLP:journals/corr/HoE16}, and FAIRL as a function the log density ratio; Figure \ref{fig:kl_figs} presents these plots.
    
    As can be seen, in AIRL (reverse KL), the policy is encouraged to place more probability mass on regions where the expert puts more mass than the policy, and less where the expert puts less. On the other hand, the GAIL (Jensen-Shannon) reward \emph{only} discourages the policy from placing more mass than the expert has.
    % this encourages the policy to not visit the modes of the expert distribution too frequently.
    
    Lastly, but very interestingly, the reward structure in FAIRL (forward KL) drastically differs from the previous two scenarios and has three distinct characteristics: (1) It encourages the policy to visit regions of the $\mathcal{S} \times \mathcal{A}$ space where the expert has put \emph{slightly less} mass than the policy; (2) It does not care if the policy places a lot more mass than the expert in some regions; (3) It \emph{severely} punishes 
        %the policy for visiting 
        regions where the expert has put more mass than the policy.
    We interpret the net effect to be that the student policy covers the expert distribution from ``outwards to inwards", meaning that it begins by placing mass in low probability regions of the space and gradually moves towards the modes of the expert's state-action distribution.
    
    % FAIRL encourages the policy to visit regions of the $\mathcal{S} \times \mathcal{A}$ space where the expert has put \emph{slightly} less mass than the policy.

    % As can be seen, in the forward KL version of AIRL, if for a state-action pair the expert puts more probability mass than the policy, the policy is \emph{severely} punished. However, if for some state-action pairs the policy places a lot more mass than the expert, it almost does not matter. As a result, the policy spreads its mass. On the other hand, in the original AIRL formulation (reverse KL), the policy is \emph{always} encouraged to put less mass than the expert. These observations are in line with standard intuitions about the mode-covering/mode-seeking behaviours of the two KL divergences \citep{bishop2006pattern}.

%% file: irl_vs_bc_exps.tex
\section{Experiments}
    \subsection{Evaluating Hypotheses}
    \label{sec:basic_exp}
    In this section we seek to empirically evaluate Hypotheses \ref{hyp:1} and \ref{hyp:2}.
    % In this section we provide empirical comparisons between AIRL, FAIRL, and standard BC in the Ant and Halfcheetah environments found in Open-AI Gym \citep{brockman2016openai}.
    % \subsection{Comparing BC, AIRL and FAIRL}
    %
       % \paragraph{Evaluation Setup}
        For each of the HalfCheetah, Ant, Walker, and Hopper simulated environments \citep{brockman2016openai} we train expert policies using Soft-Actor-Critic (SAC) \citep{haarnoja2018soft}.
        Using the trained expert policies, we generated 3 sets of expert demonstrations that contained $\{4, 16, 32\}$ trajectories. Starting from a random offset, each trajectory is subsampled by a factor of 20.\footnote{This is standard protocol employed in prior adversarial methods for Max-Ent IRL \citep{DBLP:journals/corr/HoE16, DBLP:journals/corr/abs-1710-11248}} To compare the various learning-from-demonstration algorithms we train each method at each amount of expert demonstrations using 3 random seeds. For each seed, we checkpoint the model at its best validation loss, averaged on 10 test episodes, throughout training. At the end of training, the resulting checkpoints are evaluated on $50$ test episodes.
        We defer additional implementation and experimental details to Appendix \ref{ap:irl_vs_bc_exp_details}.
        
    % \paragraph{Results \& Interpretations}

        Table \ref{table:benchmark} demonstrates that both AIRL and FAIRL outperform BC by a large margin, especially in the low data regime. The fact that FAIRL outperforms BC conclusively supports that the major performance gain of IRL methods is not due to the direction of KL divergence used, but is the result of their objectives explicitly encouraging the policy to match the marginal state distribution of the expert in addition to the matching of conditional action distribution. The fact that F/AIRL obtain similar results to DAgger \citep{ross2011reduction} further supports this claim since the DAgger algorithm was designed to mitigate the state distribution mismatch problem of BC. In conclusion, our results support Hypothesis \ref{hyp:1} while not supporting --- nor declining --- Hypothesis \ref{hyp:2}.\footnote{In our experience, in the presented benchmark settings, with sufficient hyperparameter tuning AIRL and FAIRL can outperform one-another. Additional hyperparameter tuning details discussed in Appendix \ref{app:hyper_tune}.} Concurrent to our work, \citep{ke2019imitation} present evidence that imitation learning with mode-seeking divergences may be preferable. Further investigations into \ref{hyp:2} would require evaluations in domains with multi-modal expert behaviours.
        
        \input{tables/benchmark_table}

%% file: tables/benchmark_table.tex
\begin{table*}[t]
    
    %\begin{adjustwidth}{-1in}{-1in}
        \centering
        \footnotesize
        \resizebox{\columnwidth}{!}{
        \begin{tabular}{|l|c|c||c|c||c|c||c|c|}
        \hline
        \multirow{2}{*}{\textbf{Method}} & \multicolumn{2}{c|}{\textbf{Halfcheetah}} & \multicolumn{2}{c|}{\textbf{Ant}} & \multicolumn{2}{c|}{\textbf{Walker}} & \multicolumn{2}{c|}{\textbf{Hopper}} \\ \cline{2-9} 
                                         & \textbf{Det}       & \textbf{Stoch}       & \textbf{Det}   & \textbf{Stoch}   & \textbf{Det}     & \textbf{Stoch}    & \textbf{Det}     & \textbf{Stoch}    \\ \hline
        \textbf{BC}                      &      \tpm{-62}{182}    &   \tpm{-126}{218}    &   \tpm{82}{124}    &   \tpm{19}{70}    &   \tpm{1804}{1286}    &   \tpm{1293}{480}    &   \tpm{1435}{78}    & \tpm{764}{129} \\ \hline
        \textbf{AIRL}                    &      \tpm{\textbf{8043}}{\textbf{237}}    &   \tpm{\textbf{7377}}{\textbf{482}}    &   \tpm{6024}{155}    &   \tpm{4598}{65}    &   \tpm{\textbf{3979}}{\textbf{323}}    &   \tpm{\textbf{3846}}{\textbf{319}}    &   \tpm{\textbf{3393}}{\textbf{7}}    & \tpm{\textbf{2561}}{\textbf{331}} \\ \hline
        \textbf{FAIRL}                   &      \tpm{\textbf{7924}}{\textbf{318}}    &   \tpm{\textbf{7453}}{\textbf{640}}    &   \tpm{\textbf{6607}}{\textbf{139}}    &   \tpm{\textbf{5525}}{\textbf{287}}    &   \tpm{\textbf{4297}}{\textbf{71}}    &   \tpm{\textbf{4225}}{\textbf{34}}    &   \tpm{\textbf{3379}}{\textbf{10}}    & \tpm{\textbf{3061}}{\textbf{170}} \\ \hline\hline
        
        \textbf{BC}                      &      \tpm{641}{70}    &   \tpm{285}{166}    &   \tpm{258}{292}    &   \tpm{23}{69}    &   \tpm{656.4}{72}    &   \tpm{594}{37}    &   \tpm{2543}{328}    & \tpm{1673}{375} \\ \hline
        \textbf{AIRL}                    &      \tpm{\textbf{8132}}{\textbf{143}}    &   \tpm{6914}{313}    &   \tpm{\textbf{5811}}{\textbf{208}}    &   \tpm{5027}{287}    &   \tpm{4499}{68}    &   \tpm{4355}{92}    &   \tpm{\textbf{3417}}{\textbf{4}}    & \tpm{2530}{260} \\ \hline
        \textbf{FAIRL}                   &      \tpm{\textbf{8275}}{\textbf{24}}    &   \tpm{\textbf{7900}}{\textbf{25}}    &   \tpm{\textbf{6267}}{\textbf{312}}    &   \tpm{\textbf{5473}}{\textbf{73}}    &   \tpm{\textbf{4824}}{\textbf{3}}    &   \tpm{\textbf{4778}}{\textbf{13}}    &   \tpm{\textbf{3429}}{\textbf{27}}    & \tpm{\textbf{3335}}{\textbf{38}} \\ \hline\hline
        
        \textbf{BC}                      &      \tpm{872}{640}    &   \tpm{302}{288}    &   \tpm{147}{59}    &   \tpm{94}{11}    &   \tpm{726}{33}    &   \tpm{578}{25}    &   \tpm{2253}{433}    & \tpm{1135}{407} \\ \hline
        \textbf{AIRL}                    &      \tpm{\textbf{8347}}{\textbf{37}}    &   \tpm{\textbf{7061}}{\textbf{324}}    &   \tpm{5984}{58}    &   \tpm{4406}{506}    &   \tpm{4433}{166}    &   \tpm{4284}{218}    &   \tpm{\textbf{3425}}{\textbf{14}}    & \tpm{2524}{363} \\ \hline
        \textbf{FAIRL}                   &      \tpm{\textbf{8302}}{\textbf{15}}    &   \tpm{\textbf{7522}}{\textbf{406}}    &   \tpm{\textbf{6365}}{\textbf{128}}    &    \tpm{\textbf{5442}}{\textbf{147}}   &   \tpm{\textbf{4807}}{\textbf{6}}    &   \tpm{\textbf{4764}}{\textbf{28}}    &   \tpm{\textbf{3428}}{\textbf{27}}    & \tpm{\textbf{3415}}{\textbf{21}} \\ \hline \hline
        
        \textbf{DAgger} & \tpm{8418}{14} & \tpm{6646}{1209} & \tpm{6978}{11} & \tpm{6011}{201} & \tpm{4874}{34} & \tpm{4071}{1073} & \tpm{3460}{5} & \tpm{2962}{157} \\ \hline
        
        \end{tabular}
        }
        \vspace{5pt}
        \caption{
            {
                \small
                The performance of BC, AIRL, and FAIRL on a series of standard continuous control benchmarks \citep{brockman2016openai, todorov2012mujoco}. From top to bottom, the collection of rows present results for the settings where we use 4, 16, and 32 demonstrations trajectories (with subsampling factor of 20). ``Det" and ``Stoch" respectively refer to whether we evaluate using the mode of the policy's action distribution or we sample from it. The policy architectures are held constant throughout. Hyperparameters for all models were tuned using similar budgets and values in the table report the mean and standard deviation of returns for 3 random seeds. 
                %We believe that AIRL and FAIRL obtain comparable returns and differences between their performances can be resolved by further hyperparameter tuning. 
                The key trend is that in all scenarios both IRL approaches significantly outperform BC, supporting Hypothesis \ref{hyp:1}. We also observe that F/AIRL trained with 32 demonstrations obtain a similar results to DAgger trained with the same expert data.
            }
        }
        \label{table:benchmark}
   % \end{adjustwidth}
\end{table*}

%% file: state_marginal_matching.tex
\subsection{\texorpdfstring{$f$}{f}-MAX for State Marginal Matching}
    \label{sec:smm}
    The core intuition we have built in our work is that adversarial IRL approaches additionally match state marginal distributions, rather than only action distributions as in BC. We can thus trivially apply $f$-MAX formulation to state-only marginal matching\footnote{It is important to note that state-marginal matching is not an imitation learning algorithm as is it fairly simple to design scenarios where two policies have identical state marginals yet are not the same policy.}, i.e. minimizing $D_f(\targetstatemarginal || \policystatemarginal)$, by following an iterative optimization procedure similar to $f$-MAX in Equations~\ref{eq:disc} and~\ref{eq:policy},
    \begin{align}
            \max_{T_\omega} \mathds{E}_{s \sim \targetstatemarginal}\rectbra{T_\omega(s)} - \mathds{E}_{s \sim \policystatemarginal}\rectbra{f^*(T_\omega(s))}, \qquad
            \max_\pi \mathds{E}_{\tau \sim \pi} \rectbra{\sum_t f^*(T_\omega(s_t))}
        \end{align}   
    whose justification follows identically to that of $f$-MAX with the omission of actions. 
    %To our knowledge 
    This problem was also concurrently explored in \citet{lee2019efficient} as an approach for learning effective exploration strategies. 
    In this work we are instead motivated from the direction of Imitation Learning, as an alternative approach for guiding policies that removes the need for expert demonstrations.
    Crucially, unlike in traditional IL, the target distribution need not even be a realizable state-marginal distribution; rather than gathering expensive expert demonstrations, we could rely on heuristically-designed interpretable distributions which policies will try to best match.
    % Crucially, unlike in traditional IL, the target distribution may not even need to be a realizable state marginal distribution expressed in terms of numerous expensive expert demonstrations, but rather a cheaply-designed interpretable distribution that may not be realizable but $f$-MAX tries its best to match. 
    The main difference between our setup and that of \citet{lee2019efficient} is that in this work we assume access to \emph{samples} from the target state-marginal distribution, whereas \citet{lee2019efficient} operate under the setting where they have access to the target's density function.  Similar to the IRL setting, it is not necessary to use the full state either; due to prior knowledge, or simply convenience, we may wish to match distributions over features of the state. For example, in a locomotion task we can train a policy such that its distribution of $x$-$y$ coordinates has a particular desired form.

\begin{figure}[t]
            \centering
            
            \begin{subfigure}[t]{0.7\textwidth}
                \centering
                \begin{tabular}{c@{}c@{}c@{}c}
                    \includegraphics[height=1.2in]{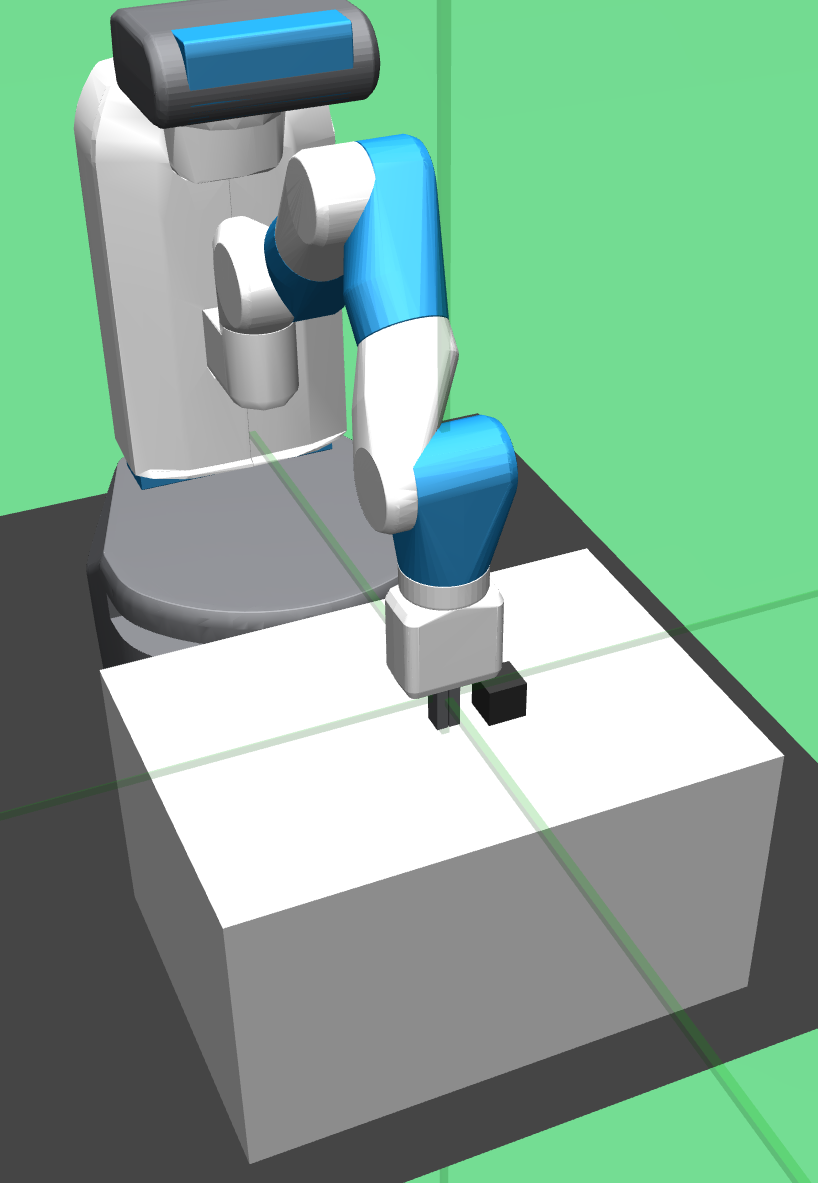}
                    & \includegraphics[height=1.2in]{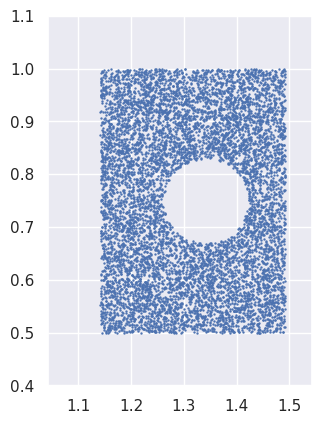}
                    & \includegraphics[height=1.2in]{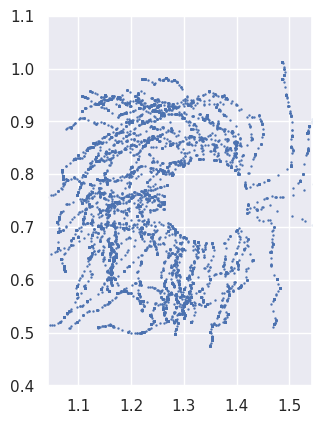}
                    % & \includegraphics[height=1.2in]{figures/slide/obj_pos_epoch_144.png}
                    & \includegraphics[height=1.2in]{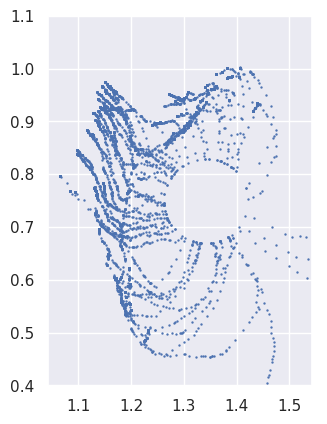}
                    % & \includegraphics[height=1.2in]{figures/slide/obj_pos_epoch_187.png}
                    % & \includegraphics[height=1.2in]{figures/slide/obj_pos_epoch_185.png}
                \end{tabular}
                \caption{Fetch Push}
                \label{fig:fetch_push}
            \end{subfigure}
            \begin{subfigure}[t]{0.29\textwidth}
                \centering
                \begin{tabular}{@{}c@{}c@{}}
                    \includegraphics[height=0.6in]{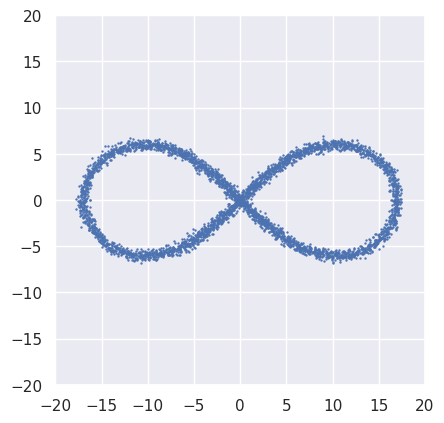} &  \includegraphics[height=0.6in]{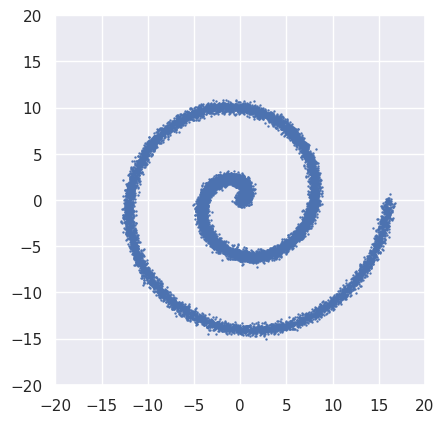}\\
                    \includegraphics[height=0.6in]{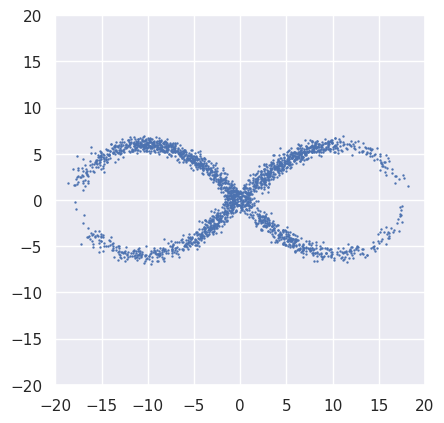} & \includegraphics[height=0.6in]{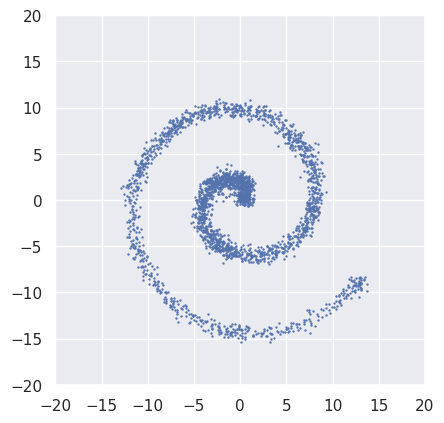}
                \end{tabular}
                \caption{Point-Mass}
                \label{fig:pm_smm}
            \end{subfigure}
            
            \begin{subfigure}[t]{0.6\textwidth}
                \centering
                \begin{tabular}{c}
                    \includegraphics[height=1.2in]{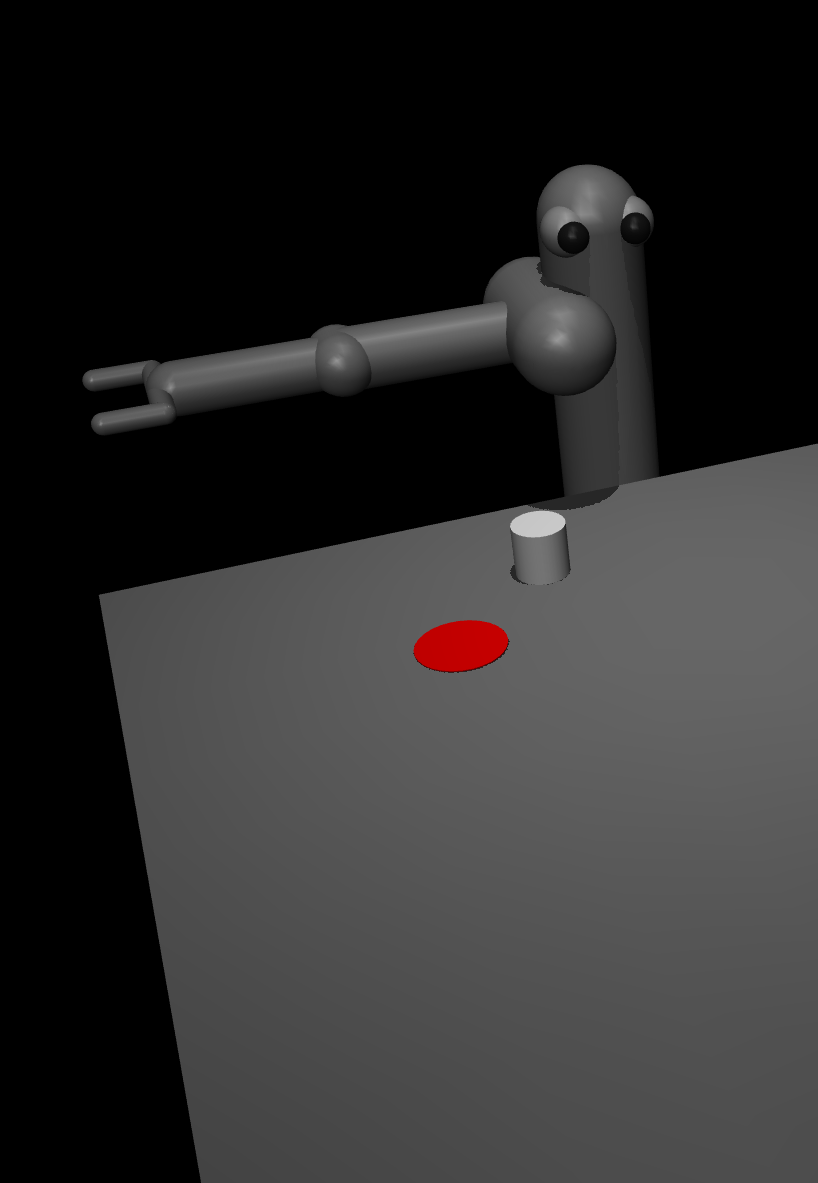}
                \end{tabular}
                % \begin{tabular}{@{}c@{}c@{\hskip6pt}c@{}c@{}}
                % % \begin{tabular}{cccc}
                %     \includegraphics[height=0.5in]{figures/pusher/pusher_obj.png} & \includegraphics[height=0.5in]{figures/pusher/pusher_arm_x_y.png} &
                %     \includegraphics[height=0.5in]{figures/pusher/pusher_policy_obj_x_y.png} & \includegraphics[height=0.5in]{figures/pusher/pusher_policy_arm_x_y.png} \\
                %     \multicolumn{2}{c}{\includegraphics[height=0.6in]{figures/pusher/pusher_arm_y_z.png}} & \multicolumn{2}{c}{\includegraphics[height=0.6in]{figures/pusher/pusher_policy_arm_y_z.png}}
                % \end{tabular}
                % \begin{tabular}{@{}c@{}c@{\hskip6pt}c@{}c@{}}
                \begin{tabular}{cc}
                % \begin{tabular}{cccc}
                    \includegraphics[height=0.5in]{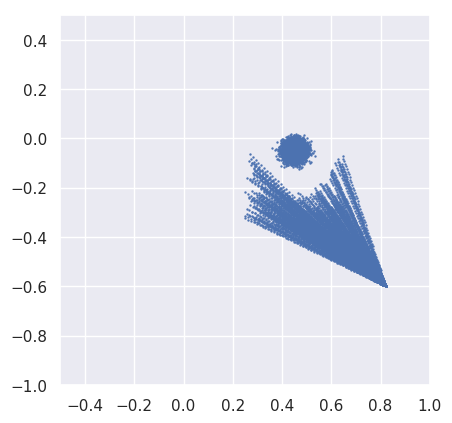} & \includegraphics[height=0.5in]{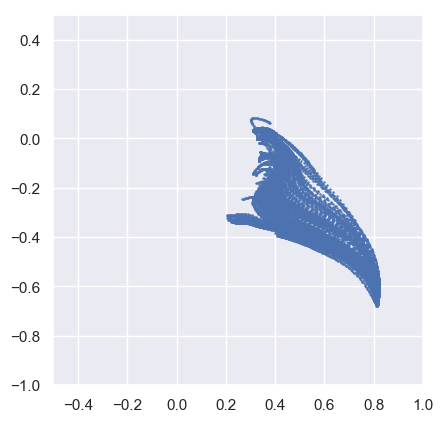} \\
                    \includegraphics[height=0.6in]{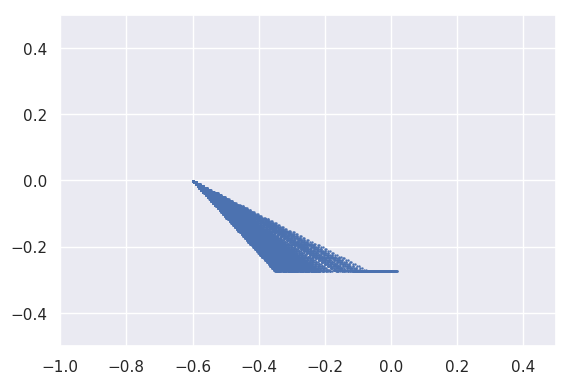} & \includegraphics[height=0.6in]{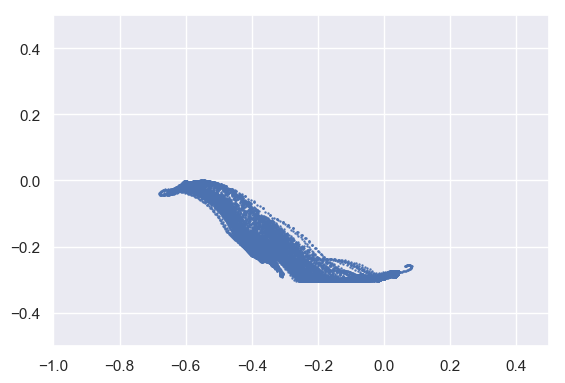}
                \end{tabular}
                \caption{Pusher Push}
                \label{fig:pusher_push}
            \end{subfigure}
            \begin{subfigure}[t]{0.39\textwidth}
                \centering
                \begin{tabular}{c@{}c}
                    \includegraphics[height=0.8in]{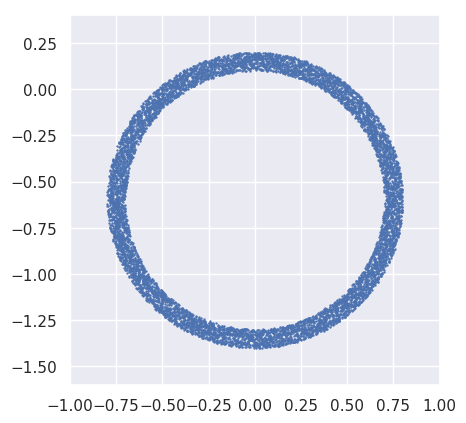} & \includegraphics[height=0.8in]{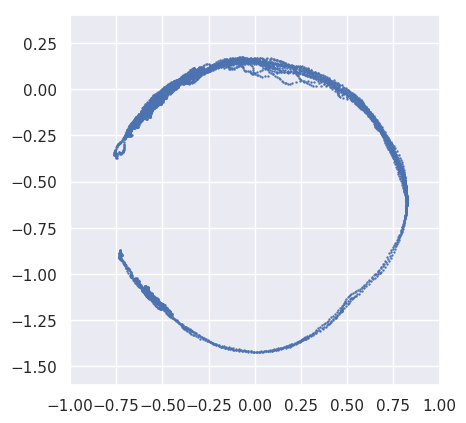}\\
                    \includegraphics[height=0.4in]{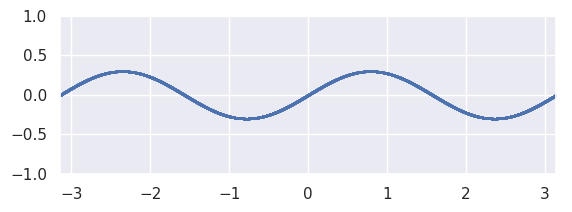} & \includegraphics[height=0.4in]{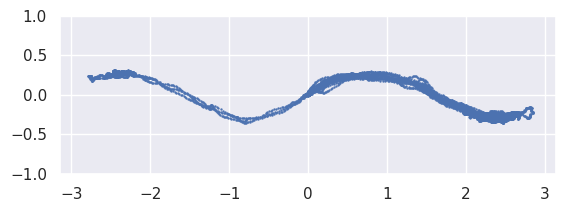}
                \end{tabular}
                \caption{Pusher Draw}
                \label{fig:pusher_draw}
            \end{subfigure}
            \caption{
                \small
                (a) Using the Fetch robot we demonstrate that we can train exploration policies through our approach to state-marginal matching. Figures in order are: Fetch environment, target, and two policies' state marginals. Full image region depicts the extent of the table. (b) In the point-mass domain we train policies that exhibit complex and multi-modal trajectories. (c) Using state-marginal matching we train policies for solving the Pusher Push task. Left image is Pusher environment.
                The next two columns correspond to the target and policy distribution of the arm tip position. Top images are bird's eye view (x-y) and bottom images are side view (y-z) of these distributions.
                % The next block of three images contains in order the object target distribution from top-down view, the arm tip target distribution from top-down view, and the arm tip target distribution from a side view. The next block of images corresponds to an example of a trained policy.
                (d) Pusher Draw target and policy distributions. Top images are top-down view of arm tip distribution and bottom images visualize the $z$ coordinate as a function of angle of rotation around the circle.
                \vspace{-0.5cm}
            }
            \label{fig:smm_results}
        \end{figure}

    \vspace{-10pt}
    \paragraph{Results}
        %In this section we will demonstrate the effectiveness of guiding policies using state-marginal distributions.
        Using various environments we design interesting target distributions and train policies using a reverse KL variant of our proposed method.
        % We will evaluate our proposed method in a simple point-mass environment and the simulated Ant environment \citep{brockman2016openai}. {\color{magenta} if time maybe we should try fetch environment too} In both environments we design a variety of target distributions on the $x$-$y$ plane with the goal of obtaining policies that learn to ``draw" such distributions on the ground.
        In the interest of space we defer experimental details to Appendix \ref{ap:state_matching_exp} and use this section to focus on main results. Videos of trained policies can be found at \url{https://sites.google.com/view/corl2019fmaxvideos/home}.
        % {
        %     \color{magenta}
        %     \begin{itemize}
        %         \item SAC, small replay buffer so it's slightly off-policy
        %         \item terminates at fixed finite horizon
        %         \item doesn't observe the time
        %     \end{itemize}
        % }
    \vspace{-10pt}    
        \paragraph{Point-Mass and Pusher Draw} Beginning with a simple point-mass domain we observe that we can train a policy to match multi-modal, complex distributions as depicted in Figure~\ref{fig:pm_smm}.
        %\paragraph{Pusher Draw} Next we move on to a similar experiment in a more complex environment. 
      Figure \ref{fig:pusher_draw} demonstrates we can also train a Pusher agent \citep{brockman2016openai} --- a joint-velocity controlled 7-DoF simulated arm --- to draw a sinusoidal function on the surface of an imaginary cylinder in 3D space.
    \vspace{-10pt}
        \paragraph{Pusher Push} Beyond path-tracing, we attempt to train the Pusher agent to solve the pushing task it was originally designed for~\citep{brockman2016openai}. The target and trained policy state distributions are presented in Figure \ref{fig:pusher_push}. %{\color{magenta} CHECK WHAT WAS THE PERFORMANCE OF THE POLICIES} 
        While this result is far from what can be achieved using RL algorithms and the provided reward function for this task, we believe this an intriguing result as our policies were merely guided by lines and points drawn in 3D space using a short python script.
        %70 line python script.
    \vspace{-10pt}
        \paragraph{Fetch Push} Lastly, in the spirit of the original motivation of \citep{lee2019efficient} we examine whether our method can lead to successful exploration strategies. To this end we use the position-controlled Fetch robot in the pick and place environment~\citep{brockman2016openai}. We fix the grippers height slightly above the table and train policies to uniformly explore the region depicted in Figure \ref{fig:fetch_push}. We observe that our training procedure gives rise to a diverse range of policies which learn to push the block around the target region, yet exhibit sufficient control to prevent the block from moving outside the boundary.
        
%% file: conclusion.tex
\section{Conclusion}

    % THE CONCLUSION FROM THE WORKSHOP PAPER
    The central motivation for this work stemmed from the superior performance of recent adversarial IRL methods \citep{DBLP:journals/corr/HoE16, DBLP:journals/corr/abs-1710-11248} compared to BC in the low-data regime, and the desire to understand the relation among various approaches for Imitation Learning. We presented $f$-MAX, a generalization of AIRL \citep{DBLP:journals/corr/abs-1710-11248} based on $f$-divergence~\citep{lin1991divergence}.
    % This unified view of imitation learning allowed us to derive a novel family of IRL algorithms and interpret AIRL, GAIL~\citep{DBLP:journals/corr/HoE16}, BC and other imitation learning algorithms as different forms of divergence minimization.
    This enabled us to form a unified view of Imitation Learning and interpret various IL methods as different forms of divergence minimization.
    From these findings, we generated hypotheses for why IRL methods outperformed BC, and empirically evaluated them in high-dimensional continuous control benchmarks. To tease apart the differences between prior IRL methods and BC, we addressed the degeneracy of $f$-MAX in a special case, and provided a one-line modification of AIRL, named FAIRL, which optimizes the forward KL divergence. Our experiments conclusively disambiguated that the factor contributing most to IRL's gain over BC is the additional state marginal matching objective.
    Lastly, we demonstrated the efficacy of applying $f$-MAX to the problem of state-marginal matching, suggesting future directions where we could replace the need for expert demonstrations in IRL with simple hand-designed state distributions.

%% file: appendix.tex
\section{Some Useful Identities}
    \label{ap:deriv}
    % We make the assumption that all epsiodes in the expert demonstrations have the same length.
    
    % \begin{align}
    %     \mathds{E}_{\tau \sim \pi} \rectbra{\sum_t f^*(T_\omega(s_t, a_t))} &= \sum_t \mathds{E}_{(s_t,a_t) \sim \polmargt} \rectbra{f^*(T_\omega(s_t, a_t))}\\
    %     &= \sum_t \int_{S,A} \polmargt f^*(T_\omega(s_t, a_t)) \\
    %     &= \int_{S,A} \rectbra{\sum_t \polmargt} f^*(T_\omega(s_t, a_t))\\
    %     &= T \cdot \int_{S,A} \polmarg f^*(T_\omega(s, a))\\
    %     &= T \cdot \mathds{E}_{(s,a) \sim \polmarg}\rectbra{f^*(T_\omega(s,a))}
    % \end{align}
    
    Let $h: \mathcal{S}\times\mathcal{A} \rightarrow \mathds{R}$ be an arbitrary function. If all episodes have the same length $T$, we have,
    
    \begin{align}
        \mathds{E}_{\tau \sim \pi} \rectbra{\sum_t h(s_t, a_t)} &= \sum_t \mathds{E}_{(s_t,a_t) \sim \polmargt} \rectbra{h(s_t, a_t)}\\
        &= \sum_t \int_{S,A} \polmargt h(s_t, a_t) \\
        &= \int_{S,A} \rectbra{\sum_t \polmargt} h(s, a)\\
        &= T \cdot \int_{S,A} \polmarg h(s, a)\\
        &= T \cdot \mathds{E}_{(s,a) \sim \polmarg}\rectbra{h(s,a)}
    \end{align}
    
    In a somewhat similar fashion, in the infinite horizon case with fixed probability $\gamma \in (0,1)$ of transitioning to a terminal state, for the discounted sum below we have,
    
    \begin{align}
        \mathds{E}_{\tau \sim \pi} \rectbra{\sum_t \gamma^t h(s_t, a_t)} &= \sum_t \mathds{E}_{(s_t,a_t) \sim \polmargt} \rectbra{\gamma^t h(s_t, a_t)}\\
        &= \sum_t \int_{S,A} \gamma^t \polmargt h(s_t, a_t) \label{eq:s1}\\
        &= \int_{S,A} \rectbra{\sum_t \gamma^t \polmargt} h(s, a) \label{eq:s2}\\
        &= \Gamma \cdot \int_{S,A} \polmarg h(s, a)\\
        &= \Gamma \cdot \mathds{E}_{(s,a) \sim \polmarg}\rectbra{h(s,a)}
    \end{align}
    
    where $\Gamma := \frac{1}{1 - \gamma}$ is the normalizer of the sum $\sum_t \gamma^t$. Since the integral of an infinite series is not always equal to the infinite series of integrals, some analytic considerations must be made to go from equation \ref{eq:s1} to \ref{eq:s2}. But, one simple case in which it holds is when the ranges of $h$ and all $\polmargt$ are bounded.

\section{More on \texorpdfstring{$f$}{f}-divergences and f-GAN}
% \paragraph{$f$-GAN}
        \label{app:f_div}
        %%(done) SG.03.19: this section comes out very randomly. add f-divergence ref in intro as suggested above + write 1-2 sentence here discussing how f-divergence relates to MaxEntIRL or GAIL. "As discussed <...>(make sure you discuss how maxent irl can be seen as minimizing KL), inverse RL can be seen as minimizing some form of divergence between \rho^exp and \rho^pi. In this section, we discuss f-divergence, which ....
        
        % \cite{DBLP:journals/corr/HoE16} demonstrate that Max-Ent IRL is the dual problem of matching $\polmarg$ to $\expmarg$; indeed as noted above, GAIL \citep{DBLP:journals/corr/HoE16} optimizes the Jensen-Shannon divergence between the two distributions. In section \ref{sec:fax} we present $f$-MAX, a method for matching $\polmarg$ to $\expmarg$ using arbitrary $f$-divergences \citep{lin1991divergence}. Hence, in this section we recall this class of statistical divergences as well as methods for using them for training generative models.
        
        Let $P, Q$ be two distributions with density functions $p, q$. For any convex, lower-semicontinuous function  $f: \mathds{R}^{+} \rightarrow \mathds{R}$ a statistical divergence can be defined as: $D_f(P||Q) = \int_\chi q(x) f\left(\frac{p(x)}{q(x)}\right)$. Divergences derived in this manner are called \emph{f-divergences} and amongst many interesting divergences include the forward and reverse KL.
        \citet{nguyen2010estimating} present a variational estimation method for $f$-divergences between arbitrary distributions P, Q. Using the notation of \citet{nowozin2016f} we can write,
            % \begin{align}
                $D_f(P||Q) \ge \sup_{T_\omega \in \mathcal{T}}(\mathds{E}_{x \sim P}\rectbra{T_\omega(x)} - \mathds{E}_{x \sim Q}\rectbra{f^*(T_\omega(x))}) \label{eq:variational_f_div}$,
            % \end{align}
        where $\mathcal{T}$ is an arbitrary class of functions $T_\omega: X \rightarrow \mathds{R}$, and $f^*$ is the convex conjugate of $f$. Under mild conditions equality holds between the two sides~\citep{nguyen2010estimating}.
        %%SG.03.19: can you just write "we have the equality" and no eq below?
            % \begin{align}
            %     D_f(P||Q) = \sup_{T_\omega \in \mathcal{T}}(\mathds{E}_{x \sim P}\rectbra{T_\omega(x)} - \mathds{E}_{x \sim Q}\rectbra{f^*(T_\omega(x))})
            % \end{align}
        Motivated by this variational approximation as well as Generative Adversarial Networks (GANs) \citep{goodfellow2014generative}, \citet{nowozin2016f} present an iterative optimization scheme for matching an implicit distribution\footnote{We use the term ``implicit distributions" to refer to distributions we can efficiently sample from (e.g. GAN \citep{goodfellow2016deep} generators) but do not necessarily have the densities of} Q to a fixed distribution P using any $f$-divergence. For a given $f$-divergence, the corresponding minimax optimization is,
        \begin{align}
            \min_Q \max_{T_\omega} \mathds{E}_{x \sim P}\rectbra{T_\omega(x)} - \mathds{E}_{x \sim Q}\rectbra{f^*(T_\omega(x))}
            % \label{eq:fgan}
        \end{align}
%        \citet{nowozin2016f} discuss practical parameterizations of $T_\omega$, but to avoid notational clutter we will use the form above.

\section{Corollary: A Simple Derivation and Intuition for AIRL}
        \label{app:airl_is_rev_kl}
        Choosing $f(u) := -\LOG u$ leads to $D_f(\expmarg || \polmarg) = \myKL{\polmarg}{\expmarg}$. This divergence is commonly referred to as the ``reverse" KL divergence. In this setting we have, $f^*(t) = -1 - \LOG(-t)$, and $T_\omega^\pi(s,a) = -\frac{\polmarg}{\expmarg}$ \citep{nowozin2016f}. Hence, given $T_\omega^\pi$, the policy objective in equation \ref{eq:policy} takes the form,
        
            \begin{align}
                &\max_\pi \mathds{E}_{\tau \sim \pi} \rectbra{\sum_t f^*(T^\pi_\omega(s_t, a_t))}\nonumber\\&= \max_\pi \mathds{E}_{\tau \sim \pi} \rectbra{\sum_t \LOG \expmargt - \LOG \polmargt - 1} \label{eq:airlform}
            \end{align}
        
        On the other hand, plugging the optimal discriminator $D^\pi(s,a) = \frac{\expmarg}{\expmarg + \polmarg}$ \citep{goodfellow2014generative} into the AIRL \citep{DBLP:journals/corr/abs-1710-11248} policy objective, we get,
        %  if in the AIRL \citep{DBLP:journals/corr/abs-1710-11248} objective we plug 
            \begin{align}
                &\max_\pi \mathds{E}_{\tau \sim \pi} \rectbra{\sum_t \LOG D^\pi(s_t,a_t) - \LOG (1-D^\pi(s_t,a_t))}\nonumber\\
                &=
                % \mathds{E}_{\tau \sim \pi} \Bigg[\sum_t \LOG \frac{\expmargt}{\expmargt + \polmargt} \nonumber\\
                % &\qquad\qquad\qquad - \LOG \frac{\polmargt}{\expmargt + \polmargt} \Bigg] \label{eq:subwithoptairl}\\
                \mathds{E}_{\tau \sim \pi} \rectbra{\sum_t \LOG \expmargt - \LOG \polmargt} \label{eq:sameasairl}
            \end{align}
            
        % where in equation \ref{eq:subwithoptairl} we replace the discriminator with its optimal form. 
        As can be seen, the right hand side of equation \ref{eq:sameasairl} matches that of equation \ref{eq:airlform} up to a constant \footnote{In both settings of fixed finite horizon, and infinite horizon with constant probability of termination, the additional term resulting from the $-1$ is a constant.}, meaning that AIRL is solving the Max-Ent IRL problem by minimizing the reverse KL divergence, $\myKL{\polmarg}{\expmarg}$!
        % $\textnormal{KL}(\polmarg||\expmarg)$, the policy has the same training objective as it does in AIRL (Equation \ref{eq:airl_kl})\footnote{If the discriminator is trained to optimality in each inner loop}.

\section{Simple Algebraic Manipulation}
    \label{ap:gail_relation}
    
    For our proof we will operate in the finite state-action space, as in the original work \citep{DBLP:journals/corr/HoE16}. In this setting, cost functions can be represented as vectors in $\mathds{R}^{\mathcal{S} \times \mathcal{A}}$, and joint state-action distributions can be represented as vectors in $[0,1]^{\mathcal{S} \times \mathcal{A}}$. Let $f$ be the function defining some $f$-divergence. Given the expert for the task, we can define the following cost function regularizer,
            \begin{align}
                \psi_f(c) := \mathds{E}_{\expmarg}\rectbra{f^*(c(s,a)) - c(s,a)}
                % = \sum_{\mathcal{S} \times \mathcal{A}} \expmarg \cdot \parbra{f^*(c(s,a)) - c(s,a)}
            \end{align}
        where $f^*$ is the convex conjugate of $f$
        % \footnote{
        %     % We must check that under this definition $\psi(c)$ is closed, proper, and convex. Convexity is simple to verify since $f^*$ is a convex function and $\expmarg > 0$.
        %     % At the moment I am not sure how to check the other two conditions.
        %     % If $f$ is proper, $f^*$ will be proper as well. 
        % }
        . Given this choice, with simple algebraic manipulation done below in Section \ref{app:alg_manipulation} we have,
        \begin{align}
            &\psi^*_f(\polmarg - \expmarg) = D_f \parbra{\polmarg || \expmarg}\\
            &\textnormal{RL} \circ \textnormal{IRL}_\psi (\pi^{\textnormal{exp}}) = \argmin_\pi -\Hcausal(\pi) + D_f \parbra{\polmarg || \expmarg}
        \end{align}
        
        Typically, the causal entropy term is considered a policy regularizer, and is weighted by $0 \le \lambda \le 1$. Therefore, modulo the term $\Hcausal(\pi)$, our derivations show that $f$-MAX, and by inheritance AIRL \citep{DBLP:journals/corr/abs-1710-11248}, all fall under the cost-regularized Max-Ent IRL framework of \cite{DBLP:journals/corr/HoE16}!
        % Furthermore, this demonstrates that the reward function of the expert can be recovered as $-T^*_\omega(s,a)$.
    
    \subsection{Algebraic Manipulation}
        \label{app:alg_manipulation}
        \begin{align}
            &\psi^*_f(\polmarg - \expmarg)\\
            &= \sup_{c \in \mathds{R}^{\mathcal{S} \times \mathcal{A}}} \rectbra{ \parbra{\polmarg - \expmarg}^T c \quad-\quad \psi_f(c)}\\
            &= \sup_{c \in \mathds{R}^{\mathcal{S} \times \mathcal{A}}} \Bigg[\sum_{\mathcal{S} \times \mathcal{A}} \parbra{\polmarg - \expmarg} \cdot c(s,a)\\
            &\qquad-\sum_{\mathcal{S} \times \mathcal{A}} \expmarg \cdot \parbra{f^*(c(s,a)) - c(s,a)}\Bigg]\\
            &= \sup_{c \in \mathds{R}^{\mathcal{S} \times \mathcal{A}}} \rectbra{\sum_{\mathcal{S} \times \mathcal{A}} \rectbra{\polmarg \cdot c(s,a) - \expmarg \cdot f^*(c(s,a))}}\\
            &= \sup_{c \in \mathds{R}^{\mathcal{S} \times \mathcal{A}}} \rectbra{\mathds{E}_{\polmarg}\rectbra{c(s,a)} - \mathds{E}_{\expmarg}\rectbra{f^*(c(s,a))}} \label{eq:with_c}\\
            &= \sup_{T_\omega \in \mathds{R}^{\mathcal{S} \times \mathcal{A}}} \rectbra{\mathds{E}_{\polmarg}\rectbra{T_\omega(s,a)} - \mathds{E}_{\expmarg}\rectbra{f^*(T_\omega(s,a))}} \label{eq:with_T}\\
            &= D_f \parbra{\polmarg || \expmarg} \label{eq:woah_gail}
        \end{align}
        To go from \ref{eq:with_c} to \ref{eq:with_T} we simply changed notation $T_\omega(s,a) := c(s,a)$, and we can go from \ref{eq:with_T} to \ref{eq:woah_gail} because it is the exact same form as the variational characterization of $f$-divergences shown in equation \ref{eq:variational_f_div}. Note that equation \ref{eq:with_T} suggests the same training procedure as described for $f$-MAX.

\section{The Problem with Forward KL}
        \label{ap:forward_KL_problem}
        Let $T^\pi_\omega$ denote the maximizer of equation \ref{eq:disc} for a given policy $\pi$. For the case of forward KL, drawing upon equations from \cite{nowozin2016f} we have,
        
        \begin{align}
            &u := \frac{\expmarg}{\polmarg}\qquad
            f(u) := u \LOG u\qquad \nonumber\\
            &f^*(t) = \textnormal{exp}(t-1)\qquad
            T_\omega^\pi = 1 + \LOG \frac{\expmarg}{\polmarg}\qquad
        \end{align}
        
        Given this, the objective for the policy (equation \ref{eq:policy}) under the optimal $T_\omega^\pi$ becomes,
        
        \begin{align}
            &\max_\pi \mathds{E}_{\tau \sim \pi} \rectbra{\sum_t f^*(T_\omega^\pi(s_t, a_t))}\\
            &\propto \mathds{E}_{(s,a) \sim \polmarg}\rectbra{f^*(T_\omega^\pi(s,a))}\\
            &= \mathds{E}_{(s,a) \sim \polmarg}\rectbra{\textnormal{exp}\left(\left(1 + \LOG \frac{\expmarg}{\polmarg}\right) - 1\right)}\\
            &= \mathds{E}_{(s,a) \sim \polmarg}\rectbra{\frac{\expmarg}{\polmarg}}\\
            &= 1
        \end{align}
        
        Hence, there is no signal to train the policy! \footnote{A similar results holds for the standard $f$-GAN formulation \citep{nowozin2016f}.}
        %%(done) SG.03.19: remove the sentence below
        % At the moment it is not clear to us how they trained their method for the case of forward KL.}!

\section{Derivation for FAIRL}
    \label{ap:FAIRL_deriv}
    % Below we present the derivation for equation \ref{eq:FAIRL_statement}. Recalling definitions,
    Below we present the derivation for FAIRL. Recalling definitions,
    \begin{align}
        &h(s,a) := \LOG D(s,a) - \LOG (1 - D(s,a))\\
        &r(s,a) := \textnormal{exp}(h(s,a)) \cdot (-h(s,a))
    \end{align}
    and assuming the discriminator is optimal\footnote{As a reminder, the optimal discriminator has the form, $D(s,a) = \frac{\expmarg}{\expmarg + \polmarg}$. A simple proof of which can be found in \cite{goodfellow2014generative}.}, we have,
    \begin{align}
        \mathds{E}_{\tau \sim \pi} \rectbra{\sum_t r(s_t, a_t)} &= \mathds{E}_{\tau \sim \pi} \rectbra{\sum_t \textnormal{exp}(h(s_t,a_t)) \cdot (-h(s_t,a_t))}\\
        &= \mathds{E}_{\tau \sim \pi} \rectbra{\sum_t \frac{\expmargt}{\polmargt} \cdot \LOG \frac{\polmargt}{\expmargt}}\\
        &\propto \mathds{E}_{(s,a) \sim \polmarg}\rectbra{\frac{\expmargt}{\polmargt} \cdot \LOG \frac{\polmargt}{\expmargt}}\\
        &= \mathds{E}_{(s,a) \sim \expmarg}\rectbra{\LOG \frac{\polmargt}{\expmargt}}\\
        &= -\textnormal{KL}(\expmarg||\polmarg)
    \end{align}

\section{Experimental Details for Hypotheses Evaluation}
    \label{ap:irl_vs_bc_exp_details}
    \paragraph{Expert Policy}
    To simulate access to expert demonstrations we train an expert policy using Soft-Actor-Critic (SAC) \citep{haarnoja2018soft}, a state-of-the-art reinforcement learning algorithm for continuous control. The expert policy consists of a 2-layer MLP with 256-dim layers, ReLU activations, and two output streams for the mean and the diagonal covariance of a \texttt{Tanh(Normal($\mu, \sigma$))} distribution \footnote{This is the architecture presented in SAC \citep{haarnoja2018soft}}. We use the default hyperparameter settings for training the expert.
    
    \paragraph{Evaluation Setup}
    Using a trained expert policy, we generated 3 sets of expert demonstrations of that contain $\{4, 16, 32\}$ trajectories. Starting from a random offset, each trajectory is subsampled by a factor of 20. This is standard protocol employed in prior direct methods for Max-Ent IRL \citep{DBLP:journals/corr/HoE16, DBLP:journals/corr/abs-1710-11248}. Also note that when generating demonstrations we \emph{sample} from the expert's action distribution rather than taking the mode. This way, since the expert was trained using Soft-Actor-Critic, the expert should correspond to the Max-Ent optimal policy for the reward function $\frac{1}{\tau} r_g(s,a)$, where $\tau$ is the SAC temperature used and $r_g(s,a)$ is the ground-truth reward function. To compare the various learning-from-demonstration algorithms we train each method at each amount of expert demonstrations using 3 random seeds. For each seed, we checkpoint the model at its best validation loss\footnote{Average return on 10 test episodes} throughout training. At the end of training, the resulting checkpoints are evaluated on $50$ test episodes.
    
    \paragraph{Details for AIRL \& FAIRL}
    For AIRL and FAIRL, the student policy has an identical architecture to that of the expert, and the discriminator is a 2-layer MLP with 256-dim layers and Tanh activations. The logits of the discriminator is clipped to be with the range \texttt{[-10,10]}. Gradient penalty is also used in the discriminator \citep{gulrajani2017improved}. We normalize the observations from the environment by computing the mean and standard deviations of the expert demonstrations. The RL algorithm used for the student policies is SAC \citep{haarnoja2018soft}, and the temperature parameter is tuned separately for AIRL \& FAIRL. We believe the combination of clipped discriminator logits, and tuning the gradient penalty as well as the SAC temperature allowed for effective training of IRL policies.
    
    \paragraph{Details for BC}
    For BC, we use an identical architecture as the expert. The model was fit using Maximum Likelihood Estimation\footnote{Recall that given a state, the output of the policy is a \texttt{Tanh(Normal($\mu, \sigma$))} distribution}. As before, the observations from the environment are normalized using the mean and standard deviation of the expert demonstrations.
    
    \paragraph{Details for DAgger}
    The DAgger algorithm was trained for 200 epochs. At the end of each epoch the policy at that point is run for 8000 timesteps. The states observed through these rollouts are labelled by the expert policy and added to the aggregate dataset.

\section{State-Marginal Matching Experimental Details}
    \label{ap:state_matching_exp}
    \subsection{Environments}
        \paragraph{Point-Mass}
        The point-mass environment we experiment with is a very simple environment implemented in NumPy with 2-dimensional observation and action space. Each episode starts with the agent initialized near the origin. At each timestep, the agent's action is a 2-dimensional vector (with magnitude of at most 1) and the agent's location is displaced by that vector. For Inifinity sign task the horizon is 120 timesteps, and for the Spiral task the horizon is 480 steps.
        
        % \paragraph{Ant}
        % {
        %     \color{magenta}
        %     For this experiment we use the Ant environment found in OpenAI gym \citep{brockman2016openai}.
        % }
        
        \paragraph{Pusher Draw}
        For this experiment we use the Pusher environment found in OpenAI gym \citep{brockman2016openai} and remove the table and object. The agent is controlled by setting velocities at various joints. We also increase the range of motion of the central joint to increase range of motion. Episode horizon is 500 timesteps.
        
        \paragraph{Pusher Push}
        For this experiment we use the Pusher environment as found in OpenAI gym \citep{brockman2016openai}. The agent is controlled by setting velocities at various joints. Episode horizon is the default 100 timesteps.
        
        \paragraph{Fetch Push}
        For this experiment we use the Fetch Pick and Place environment found in OpenAI gym \citep{brockman2016openai}. The Fetch robot is a position controlled robot, meaning that at each timestep, the agent outputs a bounded displacement vector for the gripper position. We modify the initial state distribution to start the gripper at about the same height as the block at the center of the table. In each episode the block is randomly initialized at \texttt{0.08} radius from the center of the table. We fix the gripper height and make the action space be only operate on the $x$-$y$ coordinate. Full model and environment files will be released with code release. Episode horizon is 200 timesteps.

    \subsection{Models}
        As a note, we likely used excessively large models and our choices were somewhat arbitrary with no attempt at tuning the architecture. It is very feasible that one can obtain similar performance with much more compact model architectures.
        
        \paragraph{Discriminator}
        In all experiments, the discriminator architecture is as follows. First the input is linearly embedded into a 128-dim vector. This hidden state then passes through 6 resnet blocks of 128-dimensions; the residual path uses batch normalization and Tanh activation. The last hidden state is then linearly embedded into a single-dimensional output, which is the logits of the discriminator. The logit is clipped to be within the range \texttt{[-10,10]}.
        
        \paragraph{Policy and SAC Models}
        For any given experiment the policy, Q, and V function had the same architecture, with the only difference being 1) the input-output dimensions, and 2) the policy outputs \texttt{Tanh(Normal($\mu, \sigma$))} given a state. All models used relu activations. The dimensionality and number of layer for each experiment were: Point-Mass (4 layers, 64 dim), Pusher (3 layers 128 dim), Fetch (3 layers, 256 dim).
    
    \subsection{Target Distributions}
        In this section we include the code (including parameters) used to generate the target distributions for the various experiments.
        
        \paragraph{Point-Mass}
        For the Point-Mass domain, target distributions are parameteric curves.
        
\begin{minted}{python}
import numpy as np

def infty(r, noise_scale, num_points):
    a = np.linspace(0.0, 2*np.pi, num=num_points, endpoint=True)
    X = r*(2**0.5)*np.cos(a) / (np.sin(a)**2 + 1)
    Y = X * np.sin(a)
    X += np.random.normal(scale=noise_scale, size=X.shape)
    Y += np.random.normal(scale=noise_scale, size=Y.shape)
    return X, Y

def spiral(num_rotations, radius, noise_scale, num_points):
    a = np.linspace(0.0, 2*np.pi*num_rotations, num=num_points, endpoint=True)
    r = np.linspace(0.0, radius, num=num_points, endpoint=True)
    X = r*np.cos(a)
    Y = r*np.sin(a)
    X += np.random.normal(scale=noise_scale, size=X.shape)
    Y += np.random.normal(scale=noise_scale, size=Y.shape)
    return X, Y

if __name__ == '__main__':
    X, Y = infty(12.0, 0.3, 4000)
    X, Y = spiral(2.0, 16.0, 0.3, 16000)
\end{minted}

    \paragraph{Pusher Draw}
    For the Pusher Draw task we ask the agent to trace a sinusoidal function on the surface of an imaginary cylinder.
    
\begin{minted}{python}
import numpy as np

def pusher_sin_trace(noise, num_points):
    center = np.array([0.0, -0.6])
    amp = 0.3
    a = np.linspace(-np.pi, np.pi, num=num_points, endpoint=False)
    Z = amp*np.sin(2*(a+np.pi))
    r = np.random.uniform(0.7, 0.8, size=num_points)
    X = r*np.cos(a) + center[0]
    Y = r*np.sin(a) + center[1]
    return np.stack([X,Y,Z], axis=1)

if __name__ == '__main__':
    pusher_points = pusher_sin_trace(0.2, 8000)
\end{minted}

    \paragraph{Pusher Push}
    In the Pusher Push task --- i.e. the original task designed in this environment~\citet{brockman2016openai} --- the agent must move its arm to the cylinder and push it to the target region. To examine whether we can guide policies using state-marginal matching we design the following target distribution over the joint arm tip $xyz$ position and the object $xy$ position (5-dimensional distribution). For a desired number of imaginary episodes we sample a position to initialize the cylinder at. We then sample points along the long connecting the initial arm tip coordinates to the object. Additionally, to encourage the policy to push the object to the target region. We sample many points from a Gaussian centered at the goal position.
    
\begin{minted}{python}
import numpy as np

def pusher_to_obj_to_goal_gaussian_target(num_episodes):
    obj_Z = -0.275
    init_arm_pos = np.array([8.20999983e-01, -5.99903808e-01, -1.25506088e-04])
    target_pos = np.array([0.45, -0.05, obj_Z])

    all_samples = []
    for _ in range(num_episodes):
        arm_traj = []
        obj_traj = []

        init_obj_pos = np.array([0.45, -0.05, obj_Z])
        while True:
            cylinder_pos = np.concatenate([
                np.random.uniform(low=-0.2, high=0.2, size=1),
                np.random.uniform(low=-0.3, high=0, size=1)
            ])
            if np.linalg.norm(cylinder_pos - np.zeros(2)) > 0.17:
                break
        init_obj_pos[:2] = cylinder_pos + target_pos[:2]

        # move arm to object
        w = np.linspace(0, 1, num=50, endpoint=True)[:,None]
        arm_traj.append(
            w * init_obj_pos[None,:] + (1-w) * init_arm_pos[None,:]
        )
        obj_traj.append(
            np.repeat(init_obj_pos[None,:], 50, axis=0)
        )

        # gaussian from target center
        target_region_pos = np.repeat(target_pos[None,:], 50, axis=0)
        target_region_pos[:,:2] += np.random.normal(scale=0.02, size=(target_region_pos.shape[0], 2))
        arm_traj.append(
            target_region_pos
        )
        obj_traj.append(
            target_region_pos
        )

        # put the samples together
        arm_traj = np.concatenate(arm_traj, axis=0)
        obj_traj = np.concatenate(obj_traj, axis=0)
        target_traj = np.repeat(target_pos[None,:], arm_traj.shape[0], axis=0)

        all_samples.append(
            np.concatenate([arm_traj, obj_traj[:,:2]], axis=1)
        )
    
    all_samples = np.concatenate(all_samples, axis=0)
    return all_samples

if __name__ == '__main__':
    pusher_points = pusher_to_obj_to_goal_gaussian_target(200)
\end{minted}
    
    \paragraph{Fetch Push}
    In this experiment our goal is to examine whether we can effectively explore a desired target region. To this end we draw a rectangular region contained within the table surface and uniformly sample $xy$ coordinates in this box and subtract a circular region about the initial state distribution. This forms the desired distribution over block positions. The full target distribution is a joint distribution between the gripper $xy$ coordinate and that of the object. The gripper positions is obtained by sampling from a circular region around object positions.
    
\begin{minted}{python}
import numpy as np

def uniform_box_minus_middle(num_points):
    center = np.array([1.34196849, 0.74910081])

    X = np.random.uniform(center[0] - 0.2, center[0] + 0.15, size=num_points)
    Y = np.random.uniform(center[1] - 0.25, center[1] + 0.25, size=num_points)

    obj_pos = np.stack([X, Y], axis=1)
    dist = np.linalg.norm(obj_pos - center[None, :], axis=1)
    obj_pos = obj_pos[dist > 0.08, :]

    noise = np.random.normal(size=obj_pos.shape)
    noise /= np.linalg.norm(noise, axis=1, keepdims=True)
    noise *= np.random.uniform(low=0.04, high=0.06, size=noise.shape)
    grip_pos = obj_pos + noise

    data = np.concatenate([obj_pos, grip_pos], axis=1)
    return data

if __name__ == '__main__':
    data = uniform_box_minus_middle(10000)

\end{minted}

    % \subsection{Hyperparameters}

\section{Hyperparameter Tuning}
    \label{app:hyper_tune}
    In this section we clarify our hyperparameter tuning process for AIRL and FAIRL. For both algorithms the hyperparameters to tune were 1) the Soft-Actor-Critic reward scale, and 2) the weight of the gradient penalty loss term. Our goal was to cover a wide range of benchmark domains (HalfCheetah, Ant, Walker, Hopper), with varying amounts of demonstrations (4, 16, 32), and using multiple random seeds (3 seeds). As a result, despite there being only two hyperparameters, performing a very fine hyperparameter search was unfortunately not possible for us. However, we would also like to emphasize that the gap between AIRL and FAIRL is substantially smaller than the consistent gap between BC and (A/FA)IRL.

    Our tuning process for the numbers reported in Table \ref{table:benchmark} was as follows. In initial experiments with the HalfCheetah domain we discovered that the optimal range of hyperparameters for AIRL and FAIRL differed greatly. As a result, for each domain (HalfCheetah, Ant, Walker, Hopper), at each amount of expert demonstrations (4, 16, 32), we performed a 4x4 grid search using 3 random seeds. The search grid for AIRL was \{\texttt{reward\_scale}: [2.0, 4.0, 8.0, 16.0], \texttt{grad\_pen\_weight}: [2.0, 4.0, 8.0, 16.0]\}, and the search grid for FAIRL was \{\texttt{reward\_scale}: [64.0, 128.0, 196.0, 256.0], \texttt{grad\_pen\_weight}: [0.01, 0.05, 0.1, 0.5]\}. These grids were chosen based on initial experiments with the HalfCheetah domain. In each setting, the best hyperparameters found were used to evaluate performance.
    
\begin{figure*}[h]
        \centering
        \begin{subfigure}{0.49\textwidth}
            \centering
            \includegraphics[width=1.0\linewidth]{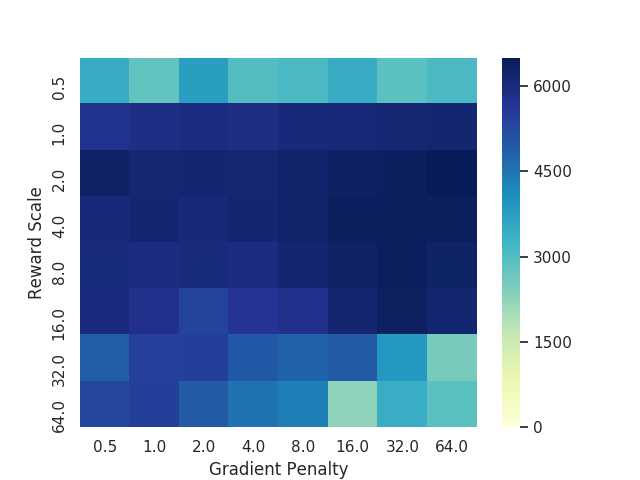}
            \caption{AIRL}
        \end{subfigure}
        \begin{subfigure}{0.49\textwidth}
            \centering
            \includegraphics[width=1.0\linewidth]{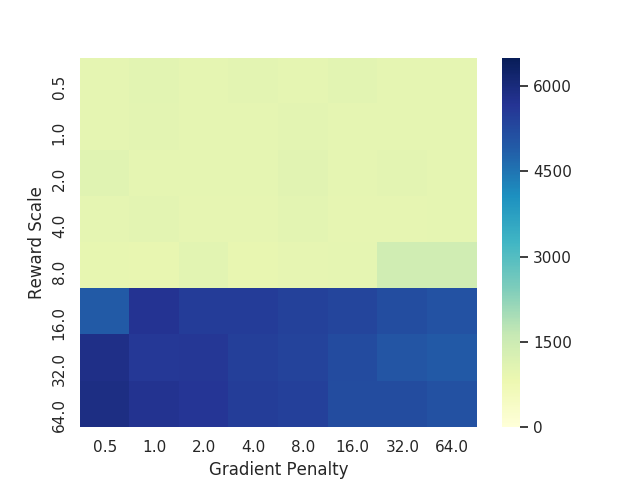}
            \caption{FAIRL}
        \end{subfigure}
        \caption{
            Hyperparameter grid search for AIRL and FAIRL in the Ant environment.
        }
        \label{fig:hype_search}
    \end{figure*}

    In our experience, it can be more challenging to find the optimal range of hyperparameters for FAIRL, likely due to the discussions of Section \ref{sec:intuition}. In the specific case of the Ant environment we performed a grid search over the two main hyperparameters: 1) the Soft-Actor-Critic reward scale, 2) The discriminator gradient penalty weight. As can be seen in \ref{fig:hype_search}, AIRL is less sensitive to these hyperparameters and obtains good performance for a wider range of hyperparameters. Note that due to the computational demand, each hyperparameter setting was run with a single random seed.
    
\section{Miscellaneous Notes}
    \paragraph{Sample-Based KL Estimates}
    Recent investigations have demonstrated that sample-based estimates of lower bounds of KL divergence may significantly under-estimate the KL divergence \citep{mcallester2018formal}. In this work we instead took the approach of estimating density ratios using a discriminator, and plugging this estimate into desired divergence. To our knowledge formal bounds on the quality of these estimates have not been discussed in the literature.
    
    \paragraph{On-Policy Training}
    While State-Marginal-Matching (SMM) can be an interesting alternative to providing expert demonstrations, similar to IRL, SMM can be quite sample-inefficient due to the need for on-policy training. Investigations into off-policy IRL and SMM methods could lead to fruitful algorithms that may be applicable in a real-world setting.
    
    \paragraph{Disconnected Modes in SMM} In SMM, if the target distribution has disconnected modes it may be very hard for a policy to discover these modes and cover the full desired target distribution. In IRL this problem does not occur since the target distribution is defined by a realizable policy (i.e. the expert policy).

%% file: ms.bbl
\begin{thebibliography}{41}
\providecommand{\natexlab}[1]{#1}
\providecommand{\url}[1]{\texttt{#1}}
\expandafter\ifx\csname urlstyle\endcsname\relax
  \providecommand{\doi}[1]{doi: #1}\else
  \providecommand{\doi}{doi: \begingroup \urlstyle{rm}\Url}\fi

\bibitem[Fu et~al.(2018)Fu, Luo, and Levine]{DBLP:journals/corr/abs-1710-11248}
J.~Fu, K.~Luo, and S.~Levine.
\newblock Learning robust rewards with adverserial inverse reinforcement
  learning.
\newblock In \emph{International Conference on Learning Representations}, 2018.
\newblock URL \url{https://openreview.net/forum?id=rkHywl-A-}.

\bibitem[Ho and Ermon(2016)]{DBLP:journals/corr/HoE16}
J.~Ho and S.~Ermon.
\newblock Generative adversarial imitation learning.
\newblock In \emph{Advances in Neural Information Processing Systems}, pages
  4565--4573, 2016.

\bibitem[Atkeson and Schaal(1997)]{atkeson1997robot}
C.~G. Atkeson and S.~Schaal.
\newblock Robot learning from demonstration.
\newblock In \emph{ICML}, volume~97, pages 12--20. Citeseer, 1997.

\bibitem[Schaal(1997)]{schaal1997learning}
S.~Schaal.
\newblock Learning from demonstration.
\newblock In \emph{Advances in neural information processing systems}, pages
  1040--1046, 1997.

\bibitem[Argall et~al.(2009)Argall, Chernova, Veloso, and
  Browning]{argall2009survey}
B.~D. Argall, S.~Chernova, M.~Veloso, and B.~Browning.
\newblock A survey of robot learning from demonstration.
\newblock \emph{Robotics and autonomous systems}, 57\penalty0 (5):\penalty0
  469--483, 2009.

\bibitem[Pastor et~al.(2009)Pastor, Hoffmann, Asfour, and
  Schaal]{pastor2009learning}
P.~Pastor, H.~Hoffmann, T.~Asfour, and S.~Schaal.
\newblock Learning and generalization of motor skills by learning from
  demonstration.
\newblock In \emph{2009 IEEE International Conference on Robotics and
  Automation}, pages 763--768. IEEE, 2009.

\bibitem[Zhu et~al.(2018)Zhu, Wang, Merel, Rusu, Erez, Cabi, Tunyasuvunakool,
  Kram{\'a}r, Hadsell, de~Freitas, et~al.]{zhu2018reinforcement}
Y.~Zhu, Z.~Wang, J.~Merel, A.~Rusu, T.~Erez, S.~Cabi, S.~Tunyasuvunakool,
  J.~Kram{\'a}r, R.~Hadsell, N.~de~Freitas, et~al.
\newblock Reinforcement and imitation learning for diverse visuomotor skills.
\newblock \emph{arXiv preprint arXiv:1802.09564}, 2018.

\bibitem[Rajeswaran et~al.(2017)Rajeswaran, Kumar, Gupta, Vezzani, Schulman,
  Todorov, and Levine]{rajeswaran2017learning}
A.~Rajeswaran, V.~Kumar, A.~Gupta, G.~Vezzani, J.~Schulman, E.~Todorov, and
  S.~Levine.
\newblock Learning complex dexterous manipulation with deep reinforcement
  learning and demonstrations.
\newblock \emph{arXiv preprint arXiv:1709.10087}, 2017.

\bibitem[Abbeel and Ng(2004)]{abbeel2004apprenticeship}
P.~Abbeel and A.~Y. Ng.
\newblock Apprenticeship learning via inverse reinforcement learning.
\newblock In \emph{Proceedings of the twenty-first international conference on
  Machine learning}, page~1. ACM, 2004.

\bibitem[Russell(1998)]{russell1998learning}
S.~J. Russell.
\newblock Learning agents for uncertain environments.
\newblock In \emph{COLT}, volume~98, pages 101--103, 1998.

\bibitem[Ng et~al.(2000)Ng, Russell, et~al.]{ng2000algorithms}
A.~Y. Ng, S.~J. Russell, et~al.
\newblock Algorithms for inverse reinforcement learning.
\newblock In \emph{Icml}, volume~1, page~2, 2000.

\bibitem[Finn et~al.(2016)Finn, Christiano, Abbeel, and
  Levine]{finn2016connection}
C.~Finn, P.~Christiano, P.~Abbeel, and S.~Levine.
\newblock A connection between generative adversarial networks, inverse
  reinforcement learning, and energy-based models.
\newblock \emph{arXiv preprint arXiv:1611.03852}, 2016.

\bibitem[Brockman et~al.(2016)Brockman, Cheung, Pettersson, Schneider,
  Schulman, Tang, and Zaremba]{brockman2016openai}
G.~Brockman, V.~Cheung, L.~Pettersson, J.~Schneider, J.~Schulman, J.~Tang, and
  W.~Zaremba.
\newblock Openai gym.
\newblock \emph{arXiv preprint arXiv:1606.01540}, 2016.

\bibitem[Lin(1991)]{lin1991divergence}
J.~Lin.
\newblock Divergence measures based on the shannon entropy.
\newblock \emph{IEEE Transactions on Information theory}, 37\penalty0
  (1):\penalty0 145--151, 1991.

\bibitem[Nowozin et~al.(2016)Nowozin, Cseke, and Tomioka]{nowozin2016f}
S.~Nowozin, B.~Cseke, and R.~Tomioka.
\newblock f-gan: Training generative neural samplers using variational
  divergence minimization.
\newblock In \emph{Advances in Neural Information Processing Systems}, pages
  271--279, 2016.

\bibitem[Lee et~al.(2019)Lee, Eysenbach, Parisotto, Xing, Levine, and
  Salakhutdinov]{lee2019efficient}
L.~Lee, B.~Eysenbach, E.~Parisotto, E.~Xing, S.~Levine, and R.~Salakhutdinov.
\newblock Efficient exploration via state marginal matching.
\newblock \emph{arXiv preprint arXiv:1906.05274}, 2019.

\bibitem[Todorov(2007)]{todorov2007linearly}
E.~Todorov.
\newblock Linearly-solvable markov decision problems.
\newblock In \emph{Advances in neural information processing systems}, pages
  1369--1376, 2007.

\bibitem[Toussaint(2009)]{toussaint2009robot}
M.~Toussaint.
\newblock Robot trajectory optimization using approximate inference.
\newblock In \emph{Proceedings of the 26th annual international conference on
  machine learning}, pages 1049--1056. ACM, 2009.

\bibitem[Peters et~al.(2010)Peters, Mulling, and Altun]{peters2010relative}
J.~Peters, K.~Mulling, and Y.~Altun.
\newblock Relative entropy policy search.
\newblock In \emph{Twenty-Fourth AAAI Conference on Artificial Intelligence},
  2010.

\bibitem[Kappen et~al.(2012)Kappen, G{\'o}mez, and Opper]{kappen2012optimal}
H.~J. Kappen, V.~G{\'o}mez, and M.~Opper.
\newblock Optimal control as a graphical model inference problem.
\newblock \emph{Machine learning}, 87\penalty0 (2):\penalty0 159--182, 2012.

\bibitem[Ziebart et~al.(2008)Ziebart, Maas, Bagnell, and
  Dey]{ziebart2008maximum}
B.~D. Ziebart, A.~L. Maas, J.~A. Bagnell, and A.~K. Dey.
\newblock Maximum entropy inverse reinforcement learning.
\newblock In \emph{Aaai}, volume~8, pages 1433--1438. Chicago, IL, USA, 2008.

\bibitem[Ziebart(2010)]{ziebart2010modeling}
B.~D. Ziebart.
\newblock Modeling purposeful adaptive behavior with the principle of maximum
  causal entropy.
\newblock 2010.

\bibitem[Haarnoja et~al.(2018)Haarnoja, Zhou, Abbeel, and
  Levine]{haarnoja2018soft}
T.~Haarnoja, A.~Zhou, P.~Abbeel, and S.~Levine.
\newblock Soft actor-critic: Off-policy maximum entropy deep reinforcement
  learning with a stochastic actor.
\newblock \emph{arXiv preprint arXiv:1801.01290}, 2018.

\bibitem[Levine(2018)]{levine2018reinforcement}
S.~Levine.
\newblock Reinforcement learning and control as probabilistic inference:
  Tutorial and review.
\newblock \emph{arXiv preprint arXiv:1805.00909}, 2018.

\bibitem[Peters and Schaal(2007)]{peters2007reinforcement}
J.~Peters and S.~Schaal.
\newblock Reinforcement learning by reward-weighted regression for operational
  space control.
\newblock In \emph{Proceedings of the 24th international conference on Machine
  learning}, pages 745--750. ACM, 2007.

\bibitem[Norouzi et~al.(2016)Norouzi, Bengio, Jaitly, Schuster, Wu, Schuurmans,
  et~al.]{norouzi2016reward}
M.~Norouzi, S.~Bengio, N.~Jaitly, M.~Schuster, Y.~Wu, D.~Schuurmans, et~al.
\newblock Reward augmented maximum likelihood for neural structured prediction.
\newblock In \emph{Advances In Neural Information Processing Systems}, pages
  1723--1731, 2016.

\bibitem[Ross et~al.(2011)Ross, Gordon, and Bagnell]{ross2011reduction}
S.~Ross, G.~Gordon, and D.~Bagnell.
\newblock A reduction of imitation learning and structured prediction to
  no-regret online learning.
\newblock In \emph{Proceedings of the fourteenth international conference on
  artificial intelligence and statistics}, pages 627--635, 2011.

\bibitem[Laskey et~al.(2017)Laskey, Lee, Hsieh, Liaw, Mahler, Fox, and
  Goldberg]{laskey2017iterative}
M.~Laskey, J.~Lee, W.~Hsieh, R.~Liaw, J.~Mahler, R.~Fox, and K.~Goldberg.
\newblock Iterative noise injection for scalable imitation learning.
\newblock In \emph{1st conference on robot learning (CoRL),(ed., Sergey Levine
  and Vincent Vanhoucke and Ken Goldberg), Mountain View, CA, USA}, pages
  13--15, 2017.

\bibitem[Coates et~al.(2008)Coates, Abbeel, and Ng]{coates2008learning}
A.~Coates, P.~Abbeel, and A.~Y. Ng.
\newblock Learning for control from multiple demonstrations.
\newblock In \emph{Proceedings of the 25th international conference on Machine
  learning}, pages 144--151. ACM, 2008.

\bibitem[Abbeel et~al.(2010)Abbeel, Coates, and Ng]{abbeel2010autonomous}
P.~Abbeel, A.~Coates, and A.~Y. Ng.
\newblock Autonomous helicopter aerobatics through apprenticeship learning.
\newblock \emph{The International Journal of Robotics Research}, 29\penalty0
  (13):\penalty0 1608--1639, 2010.

\bibitem[Dziugaite et~al.(2015)Dziugaite, Roy, and
  Ghahramani]{dziugaite2015training}
G.~K. Dziugaite, D.~M. Roy, and Z.~Ghahramani.
\newblock Training generative neural networks via maximum mean discrepancy
  optimization.
\newblock \emph{arXiv preprint arXiv:1505.03906}, 2015.

\bibitem[Li et~al.(2015)Li, Swersky, and Zemel]{li2015generative}
Y.~Li, K.~Swersky, and R.~Zemel.
\newblock Generative moment matching networks.
\newblock In \emph{International Conference on Machine Learning}, pages
  1718--1727, 2015.

\bibitem[Goodfellow et~al.(2014)Goodfellow, Pouget-Abadie, Mirza, Xu,
  Warde-Farley, Ozair, Courville, and Bengio]{goodfellow2014generative}
I.~Goodfellow, J.~Pouget-Abadie, M.~Mirza, B.~Xu, D.~Warde-Farley, S.~Ozair,
  A.~Courville, and Y.~Bengio.
\newblock Generative adversarial nets.
\newblock In \emph{Advances in neural information processing systems}, pages
  2672--2680, 2014.

\bibitem[Ke et~al.(2019)Ke, Barnes, Sun, Lee, Choudhury, and
  Srinivasa]{ke2019imitation}
L.~Ke, M.~Barnes, W.~Sun, G.~Lee, S.~Choudhury, and S.~Srinivasa.
\newblock Imitation learning as $ f $-divergence minimization.
\newblock \emph{arXiv preprint arXiv:1905.12888}, 2019.

\bibitem[Bloem and Bambos(2014)]{bloem2014infinite}
M.~Bloem and N.~Bambos.
\newblock Infinite time horizon maximum causal entropy inverse reinforcement
  learning.
\newblock In \emph{53rd IEEE Conference on Decision and Control}, pages
  4911--4916. IEEE, 2014.

\bibitem[Todorov et~al.(2012)Todorov, Erez, and Tassa]{todorov2012mujoco}
E.~Todorov, T.~Erez, and Y.~Tassa.
\newblock Mujoco: A physics engine for model-based control.
\newblock In \emph{2012 IEEE/RSJ International Conference on Intelligent Robots
  and Systems}, pages 5026--5033. IEEE, 2012.

\bibitem[Nguyen et~al.(2010)Nguyen, Wainwright, and
  Jordan]{nguyen2010estimating}
X.~Nguyen, M.~J. Wainwright, and M.~I. Jordan.
\newblock Estimating divergence functionals and the likelihood ratio by convex
  risk minimization.
\newblock \emph{IEEE Transactions on Information Theory}, 56\penalty0
  (11):\penalty0 5847--5861, 2010.

\bibitem[Goodfellow et~al.(2016)Goodfellow, Bengio, Courville, and
  Bengio]{goodfellow2016deep}
I.~Goodfellow, Y.~Bengio, A.~Courville, and Y.~Bengio.
\newblock \emph{Deep learning}, volume~1.
\newblock MIT Press, 2016.

\bibitem[Bishop(2006)]{bishop2006pattern}
C.~M. Bishop.
\newblock \emph{Pattern recognition and machine learning}.
\newblock springer, 2006.

\bibitem[Gulrajani et~al.(2017)Gulrajani, Ahmed, Arjovsky, Dumoulin, and
  Courville]{gulrajani2017improved}
I.~Gulrajani, F.~Ahmed, M.~Arjovsky, V.~Dumoulin, and A.~C. Courville.
\newblock Improved training of wasserstein gans.
\newblock In \emph{Advances in neural information processing systems}, pages
  5767--5777, 2017.

\bibitem[McAllester and Statos(2018)]{mcallester2018formal}
D.~McAllester and K.~Statos.
\newblock Formal limitations on the measurement of mutual information.
\newblock \emph{arXiv preprint arXiv:1811.04251}, 2018.

\end{thebibliography}
